\documentclass[runningheads]{llncs}

\usepackage[utf8]{inputenc} 
\usepackage[T1]{fontenc}    
\usepackage{hyperref}       
\usepackage{url}            
\usepackage{booktabs}       
\usepackage{amsfonts}       
\usepackage{nicefrac}       
\usepackage{microtype}      
\usepackage{xcolor}         
\usepackage{amsmath}
\usepackage{graphicx}
\usepackage{rotating}
\usepackage{tabulary}
\usepackage{algorithm}
\usepackage{algorithmic}
\usepackage{subdef}
\usepackage{wrapfig,lipsum,booktabs}
\usepackage{subcaption}
\usepackage[numbers]{natbib}
\usepackage[misc]{ifsym}

\newcommand{\figref}[1]{Fig.~\ref{#1}}
\newcommand{\tabref}[1]{Tab.~\ref{#1}}
\newcommand{\secref}[1]{Sec.~\ref{#1}}
\newcommand{\AlgRef}[1]{Algo.~\ref{#1}}

\newcommand{\Appref}[1]{Appendix.~\ref{#1}}
\newcommand{\model}{\mbox{\textsc{Diagnose}}}

\newcommand{\scmi}{\mbox{\textsc{Scmi}}}
\newcommand{\smi}{\mbox{\textsc{Smi}}}
\newcommand{\scg}{\mbox{\textsc{Scg}}}
\begin{document}

\title{DIAGNOSE: Avoiding Out-of-distribution Data using Submodular Information Measures}
\titlerunning{DIAGNOSE: Avoiding OOD Data using Submodular Information Measures}
%
\author{Suraj Kothawade\inst{1}$^{\textrm{\Letter}}$ \and
Akshit Srivastava\inst{2} \and
Venkat Iyer\inst{2} \and
Ganesh Ramakrishnan\inst{2}\and
Rishabh Iyer\inst{1}}
\authorrunning{S. Kothawade et al.}
%
\institute{University of Texas at Dallas, USA \and
Indian Institute of Technology, Bombay, India\\
\email{suraj.kothawade@utdallas.edu}}
\maketitle              

\begin{abstract}

Avoiding out-of-distribution (OOD) data is critical for training supervised machine learning models in the medical imaging domain. Furthermore, obtaining labeled medical data is difficult and expensive since it requires expert annotators like doctors, radiologists, \etc\ Active learning (AL) is a well-known method to mitigate labeling costs by selecting the most diverse or uncertain samples. However, current AL methods do not work well in the medical imaging domain with OOD data. We propose \model\ (avoi\textbf{D}ing out-of-d\textbf{I}stribution d\textbf{A}ta usin\textbf{G} submodular i\textbf{N}f\textbf{O}rmation mea\textbf{S}ur\textbf{E}s), an active learning framework that can jointly model similarity and dissimilarity, which is crucial in mining in-distribution data and avoiding OOD data at the same time. Particularly, we use a small number of data points as exemplars that represent a query set of in-distribution data points and another set of exemplars that represent a private set of OOD data points. We illustrate the generalizability of our framework by evaluating it on a wide variety of real-world OOD scenarios. Our experiments verify the superiority of \model\ over the state-of-the-art AL methods across multiple domains of medical imaging.

\end{abstract}

\section{Introduction}

Deep learning based models are widely used for medical image computing. However, it is critical to mitigate incorrect predictions for avoiding a catastrophe when these models are deployed at a health-care facility. It is known that deep models are data hungry, which leads us to two problems before we can train a high quality model. \textbf{Firstly,} procuring medical data is difficult due to limited availability and privacy constraints. \textbf{Secondly,} acquiring the \emph{right} labeled data to train a supervised model which has minimum dissimilarity with the test (deployment) distribution can be challenging \cite{saria2019tutorial}. This difficulty is particularly because the unlabeled dataset consists of out-of-distribution (OOD) data caused due to changes in data collection procedures, treatment protocols, demographics of the target population, \etc\ \cite{finlayson2021clinician}. In this paper, we study active learning (AL) strategies in order to mitigate \emph{both} these problems.

Current AL techniques are designed to acquire data points that are either the most uncertain, or the most diverse, or a mix of both. Unfortunately, this makes the current techniques susceptible to picking data points that are OOD which gives rise to two more problems: \textbf{1)} Wastage of expensive labeling resources, since expert annotators need to filter out OOD data points rather than focusing on annotating the in-distribution data points. \textbf{2)} Drop in model performance, since OOD data points may sink into the labeled set due to human errors. To tackle the above problems, we propose \model, an active learning framework that uses the submodular information measures \cite{iyer2020submodular} as acquisition functions to model similarity with the in-distribution data points and dissimilarity with the OOD data points. \looseness-1 

\begin{wrapfigure}{R}{0.50\textwidth}
\centering
\includegraphics[width = 0.5\textwidth, height=6cm]{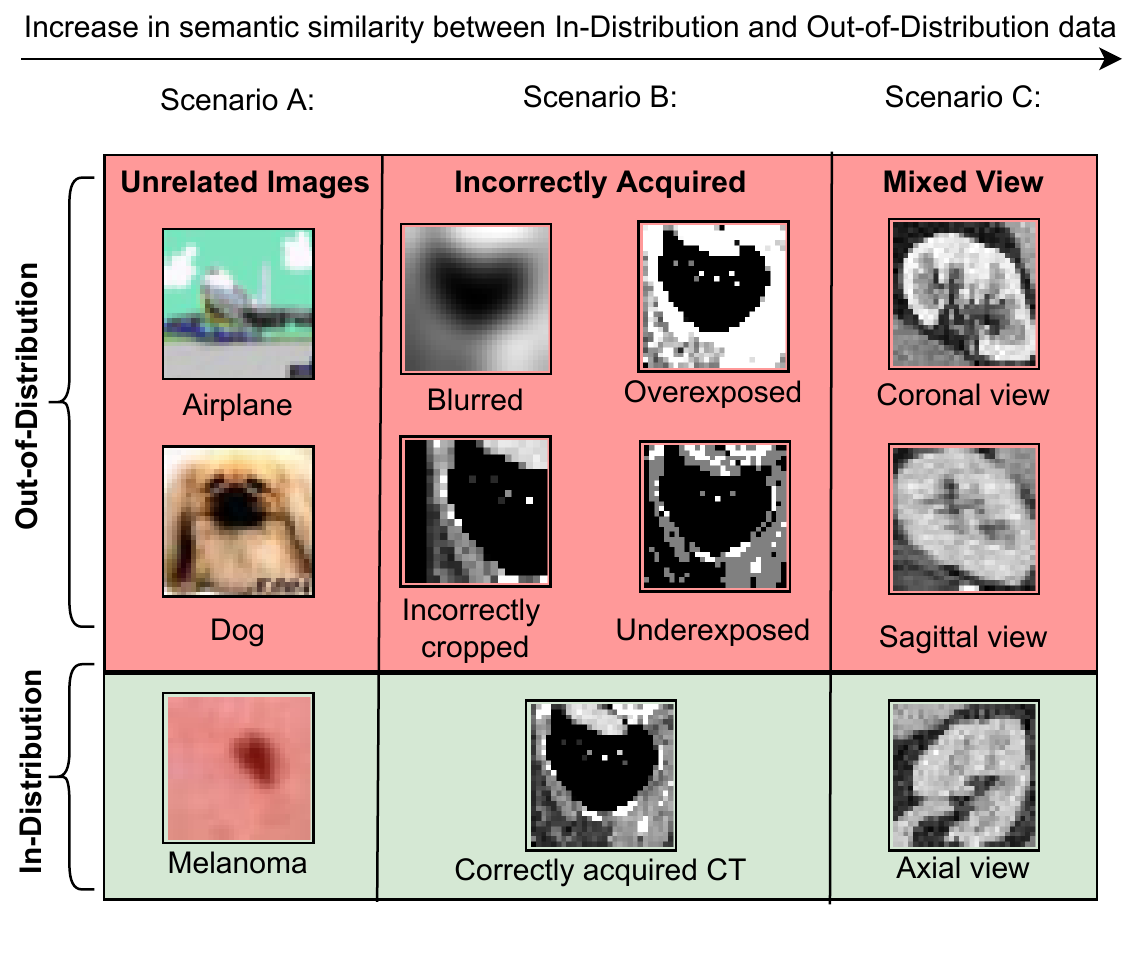}

\caption{The out-of-distribution (OOD) images in three scenarios are contrasted with the in-distribution (ID) images.       
A: Inputs that are unrelated to the task.
B: Inputs which are incorrectly acquired.
C: Inputs that belong to a different view of anatomy. Note that these scenarios become \emph{increasingly} difficult as we go from A $\rightarrow$ C since the semantic similarity between OOD and ID increases.}
\label{fig:scenarios_3}
\end{wrapfigure}
\vspace{-3ex}

\subsection{Problem Statement: OOD Scenarios in Medical Data} \label{sec:ood_scenarios}

We consider a diverse set of \emph{four} OOD scenarios with increasing levels of difficulty. We present three scenarios in \figref{fig:scenarios_3} and discuss an additional scenario in \Appref{app:scenario_D}. We present the details for each scenario in the context of image classification below:

\vspace{0.2cm}

\noindent \textbf{Scenario A - Unrelated Images:} Avoid images that are completely unassociated for the task. For instance, real-world images mixed with skin lesion images (first column in \figref{fig:scenarios_3}).

\vspace{0.2cm}

\noindent \textbf{Scenario B - Incorrectly Acquired:} Avoid images that are either captured incorrectly or post-processed incorrectly. For instance, incorrectly cropped/positioned images, blurred images, or images captured using a different procedure \etc\ (second column in \figref{fig:scenarios_3}). OOD images of this type are harder to filter than scenario A since there may be some overlap with the semantics of the in-distribution images.

\vspace{0.2cm}

\noindent \textbf{Scenario C - Mixed View:} Avoid images captured with a different view of the anatomy than the deployment scenario. For example, images from a coronal or sagittal view are OOD when the deployment is on axial view images (third column in \figref{fig:scenarios_3}). Note that this scenario is further challenging than scenario B since \emph{only} the viewpoint of the \emph{same} organ makes it ID or OOD.

\subsection{Related work} \label{sec:related_work}

\textbf{Uncertainty based Active Learning.} Uncertainty based methods aim to select the most uncertain data points according to a model for labeling. The most common techniques are - 1) \textsc{Entropy} \cite{settles2009active} selects data points with maximum entropy, and 2) \textsc{Margin} \cite{roth2006margin} selects data points such that the difference between the top two predictions is minimum.

\vspace{0.2cm}

\noindent \textbf{Diversity based Active Learning.} The main drawback of uncertainty based methods is that they lack diversity within the acquired subset. To mitigate this, a number of approaches have proposed to incorporate diversity. The \textsc{Coreset} method \cite{sener2018active} minimizes a coreset loss to form coresets that represent the geometric structure of the original dataset. They do so using a greedy \textit{k}-center clustering. A recent approach called \textsc{Badge}~\cite{ash2020deep} uses the last linear layer gradients to represent data points and runs \textsc{K-means++}~\cite{kmeansplus} to obtain centers each having high gradient magnitude. Having representative centers with high gradient magnitude ensures uncertainty and diversity at the same time. However, for batch AL, \textsc{Badge} models diversity and uncertainty only within the batch and \emph{not} across all batches. Another method, \textsc{BatchBald} \cite{kirsch2019batchbald} requires a large number of Monte Carlo dropout samples to obtain reliable mutual information which limits its application to medical domains where data is scarce.

\vspace{0.2cm}

\noindent \textbf{Active Learning for OOD data. }To the best of our knowledge, only a small minority of AL methods tackle OOD data. Our work is closest to and inspired from \textsc{Similar} \cite{kothawade2021similar}, which uses the \scmi\ functions (see \secref{sec:preliminaries}) for \emph{simulated} OOD scenarios on \emph{toy} datasets with thumbnail images (CIFAR-10 \cite{krizhevsky2009learning}) and black and white digit images (MNIST \cite{lecun2010mnist}). In contrast, \model\ tackles a wide range of real-world OOD scenarios in the medical imaging domain. Another related AL baseline is \textsc{Glister-Active} \cite{killamsetty2020glister} with an acquisition formulation that maximizes the log-likelihood on a held-out validation set.   


\subsection{Our contributions}
We summarize our contributions as follows: \textbf{1)} We emphasize on \emph{four} diverse OOD data scenarios in the context of medical image classification (see \figref{fig:scenarios_3}). \textbf{2)} Given the limitations of current AL methods on medical datasets, we propose \model, a novel AL framework that can jointly model similarity with the in-distribution (ID) data points and dissimilarity with the OOD data points. We observe that the submodular conditional mutual information functions that jointly model similarity and dissimilarity acquire the most number of ID data points (see \figref{fig:res_ood_1}, 3, 4). \textbf{3)} We demonstrate the effectiveness of our framework for multiple modalities, namely, dermatoscopy, Abdominal CT, and histopathology. Furthermore, we show that \model\ consistently outperforms the state-of-the-art AL methods on all OOD scenarios. \textbf{4)} Through rigorous ablation studies, we compare the effects of maximizing mutual information and conditional gain functions. \looseness-1 

\section{Preliminaries} \label{sec:preliminaries}

\textbf{Submodular Functions: } We let $\Vcal$ denote the \emph{ground-set} of $n$ data points $\Vcal = \{1, 2, 3,...,n \}$ and a set function $f:
 2^{\Vcal} \xrightarrow{} \mathbb{R}$.  
 The function $f$ is submodular~\citep{fujishige2005submodular}  if it satisfies the diminishing marginal returns, namely $f(j | \Acal) \geq f(j | \Bcal)$ for all $\Acal \subseteq \Bcal \subseteq \Vcal, j \notin \Bcal$.
 Different submodular functions model different properties. For \eg, facility location, $f(\Acal)=\sum\limits_{i \in \Vcal} \max\limits_{j \in \Acal} S_{ij}$, selects a representative subset and log determinant, $f(\Acal)=\log\det(S)$ selects a diverse subset~\citep{iyer2015submodular}, where $S$ is a matrix containing pariwise similarity values $S_{ij}$.
\vspace{-2ex}

\begin{table}[!htb]
    \caption{Instantiations of Submodular Information Measures (SIM).
    }
    \label{tab:SIM_inst}
    \begin{subtable}{.4\linewidth}
      \centering
        \caption{SMI and SCG functions.}
        \label{tab:smi_inst}
        \begin{tabular}{|c|c|c|}
        \hline
        \textbf{SMI} & \textbf{$I_f(\Acal;\Qcal)$} \\ \hline
        \scriptsize{FLMI}             & \scriptsize{$\sum\limits_{i \in \Ucal}\min(\max\limits_{j \in \Acal}S_{ij},  \max\limits_{j \in \Qcal}S_{ij})$}                \\
        \scriptsize{LogDetMI}          & \scriptsize{$\log\det(S_{\Acal}) -\log\det(S_{\Acal} -$} \\ & \scriptsize{$ S_{\Acal,\Qcal}S_{\Qcal}^{-1}S_{\Acal,\Qcal}^T)$}           \\ \hline
        \end{tabular}
        
        \vspace{0.5ex}
        
         \begin{tabular}{|l|l|}
            \hline
            \textbf{SCG} & \textbf{$f(\Acal|\Pcal)$} \\ \hline
            \scriptsize{FLCG}       & \scriptsize{$\sum\limits_{i \in \Ucal} \max(\max\limits_{j \in \Acal} S_{ij}-$ $ \max\limits_{j \in \Pcal} S_{ij}, 0)$}                \\
            \scriptsize{LogDetCG}       & \scriptsize{$\log\det(S_{\Acal} - S_{\Acal,\Pcal}S_{\Pcal}^{-1}S_{\Acal,\Pcal}^T)$} \\
            \hline
        \end{tabular}
    \end{subtable}%
    \begin{subtable}{.6\linewidth}
      \centering
        \caption{SCMI functions.}
        \label{tab:scg_scmi_inst}

        \begin{tabular}{|l|l|}
        \hline
        \textbf{SCMI} & \textbf{$I_f(\Acal;\Qcal|\Pcal)$} \\ \hline
        \scriptsize{FLCMI}       & \scriptsize{$\sum\limits_{i \in \Ucal} \max(\min(\max\limits_{j \in \Acal} S_{ij},$ $ \max\limits_{j \in \Qcal} S_{ij})$} \\ & \scriptsize{$-  \max\limits_{j \in \Pcal} S_{ij}, 0)$}                \\\scriptsize{
        LogDetCMI}   & \scriptsize{$\log \frac{\det(I - S_{\Pcal}^{-1}S_{\Pcal, \Qcal} S_{\Qcal}^{-1}S_{\Pcal, \Qcal}^T)}{\det(I - S_{\Acal \cup \Pcal}^{-1} S_{\Acal \cup \Pcal, Q} S_{\Qcal}^{-1} S_{\Acal \cup \Pcal, Q}^T)}$} \\ \hline               
        \end{tabular}
    \end{subtable} 
\end{table}

\noindent \textbf{Submodular Information Measures (SIM):} Given a set of items $\Acal, \Qcal, \Pcal \subseteq \Vcal$, the submodular conditional mutual information (\scmi)~\citep{iyer2020submodular} is defined as $I_f(\Acal; \Qcal | \Pcal) = f(\Acal \cup \Pcal) + f(\Qcal \cup \Pcal) - f(\Acal \cup \Qcal \cup \Pcal) - f(\Pcal)$. Intuitively, this jointly measures the similarity between $\Qcal$ and $\Acal$ and the dissimilarity between $\Pcal$ and $\Acal$. We refer to $\Qcal$ as the query set and $\Pcal$ as the private or conditioning set. Kothawade {\em et. al.}~\cite{kothawade2021prism} extend the SIM to handle the case when $\Qcal$ and $\Pcal$ can come from a different set $\Vcal'$ which is disjoint from the ground set $\Vcal$. In the context of medical image classification in scenarios with OOD data, $\Vcal$ is the source set of images, whereas $\Qcal$ contains data points from the in-distribution classes that we are interested in selecting, and $\Pcal$ contains OOD data points that we want to avoid.
As discussed in \cite{kothawade2021similar}, we can use the \scmi\ formulation to obtain the submodular mutual information (\smi) by setting $\Qcal \leftarrow \Qcal$ and $\Pcal \leftarrow \emptyset$. The \smi\ is defined as: $I_f(\Acal; \Qcal) = f(\Acal) + f(\Qcal) - f(\Acal \cup \Qcal)$. Similarly, the submodular conditional gain (\scg) formulation can be obtained by setting $\Qcal \leftarrow \emptyset$ and $\Pcal \leftarrow \Pcal$. The \scg\ is defined as: $f(\Acal | \Pcal) = f(\Acal \cup \Pcal) - f(\Pcal)$. To find an optimal subset given $\Qcal, \Pcal \subseteq \Vcal^{\prime}$, we can define $g_{\Qcal, \Pcal}(\Acal) = I_f(\Acal; \Qcal | \Pcal)$, $\Acal \subseteq \Vcal$ and maximize the same. In \tabref{tab:SIM_inst}, we present the instantiations of various \scmi, \scg\ and, \scmi\ functions with the naming convention abbreviated as the `function name' + `CMI/MI/CG'. The submodular functions that we use include `Facility Location' (FL) and `Log Determinant' (LogDet) ~\cite{iyer2020submodular, kothawade2021prism}.

\begin{figure}[h]
\centering
\includegraphics[width = \textwidth]{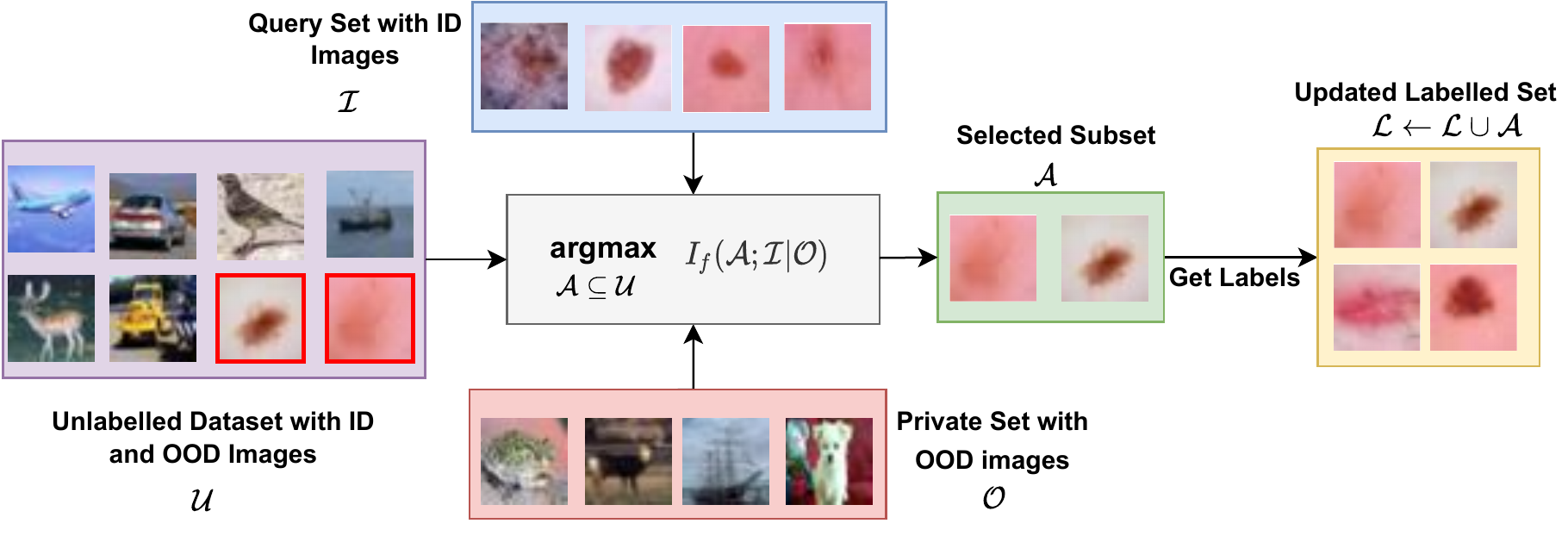}

\caption{One round of active learning using \model. We optimize the \scmi\ function to jointly model similarity with the query set $\Ical$ with ID images and dissimilarity with the private set $\Ocal$ with OOD images.}
\label{fig:flowchart}
\end{figure}

\vspace{-2ex}

\section{Leveraging Submodular Information Measures for Multiple Out-of-distribution Scenarios} \label{sec:our_method}
In this section, we present \model\ (see \figref{fig:flowchart}), a one-stop framework that uses the \scmi\ functions as AL acquistion functions to tackle OOD scenarios (\figref{fig:scenarios_3}). 
\begin{algorithm}
\begin{algorithmic}[1]
\REQUIRE Initial labeled set: $\Lcal$, Initial set of ID points: $\Ical \leftarrow \Lcal$, Initial set of OOD points: $\Ocal \leftarrow \emptyset$ large unlabeled dataset: $\Ucal$ with ID and OOD points, Loss function $\Hcal$ for learning model $\Mcal$, batch size: $B$, number of selection rounds: $N$ \\
\FOR{selection round $i = 1:N$}
\STATE Train $\Mcal_{\theta_i}$ with loss $\Hcal$ on the current labeled set $\Lcal$.
\STATE $\Gcal_\Ucal \leftarrow  \{\nabla_{\theta_i} \mathcal H(x_j, \hat{y_j}, \theta_i), \forall j \in \Ucal\}$ \textcolor{blue}{\{Compute gradients using hypothesized labels\}}
\STATE $\Gcal_\Ical, \Gcal_\Ocal \leftarrow \{\nabla_{\theta_i} \mathcal H(x_j, y_j, \theta_i), \forall j \in \Ical, \Ocal\}$ \textcolor{blue}{\{Compute gradients using true labels\}}
\STATE $\Xcal \leftarrow$ \textsc{Cosine\_Similarity} ($\{\Gcal_\Ical \cup \Gcal_\Ocal\}, \Gcal_\Ucal$) \textcolor{blue}{\{$X \in \mathbb{R}^{|\Ical \cup \Ocal| \times |\Ucal|}$\}}
\STATE Instantiate a \scmi\ function $I_f$ based on $\Xcal$.
\STATE $\Acal_i \leftarrow \mbox{argmax}_{\Acal \subseteq \Ucal, |\Acal | \leq B}  I_f(\Acal; \Ical | \Ocal)$ 
\STATE Get labels $L(\Acal_i)$ for batch $\Acal_i$ and $\Lcal \leftarrow \Lcal \cup L(\Acal_i)$, $\Ucal \leftarrow \Ucal - \Acal_i$
\STATE $\Ical \leftarrow \Ical \cup \Acal_i^\Ical$, $\Ocal \leftarrow \Ocal \cup \Acal_i^\Ocal$ \textcolor{blue}{\{Add new ID points to $\Ical$ and new OOD points to $\Ocal$\}}
\ENDFOR
\STATE Return trained model $\Mcal$ and parameters $\theta$.
\end{algorithmic}
\caption{\model: Avoiding OOD using SIM}
\label{algo:diagnose}
\end{algorithm}
The main idea in our approach is to exploit the joint modeling of similarity and dissimilarity in \scmi\ functions to acquire the desired in-distribution (ID) data and avoid the out-of-distribution (OOD) data. We do so by maintaining two sets, {\em viz.}: $\Ical$ containing the ID data points, and $\Ocal$ containing the OOD data points that we have encountered so far in the batch active learning loop. Next, we assign the query set $\Qcal \leftarrow \Ical$ and the private set $\Pcal \leftarrow \Ocal$ in the \scmi\ formulation (see \secref{sec:preliminaries}). Using last layer gradients as a representation for each data point, we compute the similarity matrix $\Xcal$ between the unlabeled set $\Ucal$ and $\{\Ical \cup \Ocal\}$. We then optimize the resulting function $I_f(\Acal;\Ical|\Ocal)$ instantiated by $\Xcal$ using a greedy strategy~\cite{mirzasoleiman2015lazier}. In any AL round $i$, we use the ID data points from newly acquired labeled set $A_i^\Ical \subseteq A_i$ to augment $\Ical \leftarrow \Ical \cup A_i^\Ical$, and new OOD data points $A_i^\Ocal \subseteq A_i$ to augment $\Ocal \leftarrow \Ocal \cup A_i^\Ocal$. Note that $\Acal_i \leftarrow \Acal_i^\Ical \cup \Acal_i^\Ocal$. In our experiments (see \secref{sec:experiments}), we also use the corresponding \smi\ formulation $I_f(\Acal;\Ical)$, and the \scg\ formulation $f(\Acal|\Ocal)$ as acquisition functions. We summarize \model\ in \AlgRef{algo:diagnose} and discuss its scalability aspects in \Appref{app:scalability}.

\section{Experimental Results} \label{sec:experiments}

\begin{figure*}[h]
\centering
\includegraphics[width = 12cm, height=0.7cm]{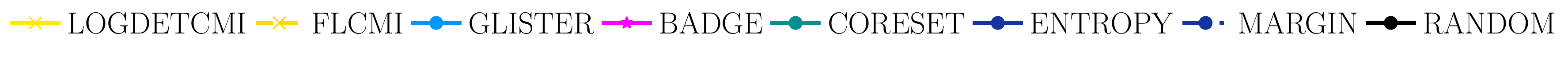}
\begin{subfigure}[]{0.33\textwidth}
\includegraphics[width = \textwidth]{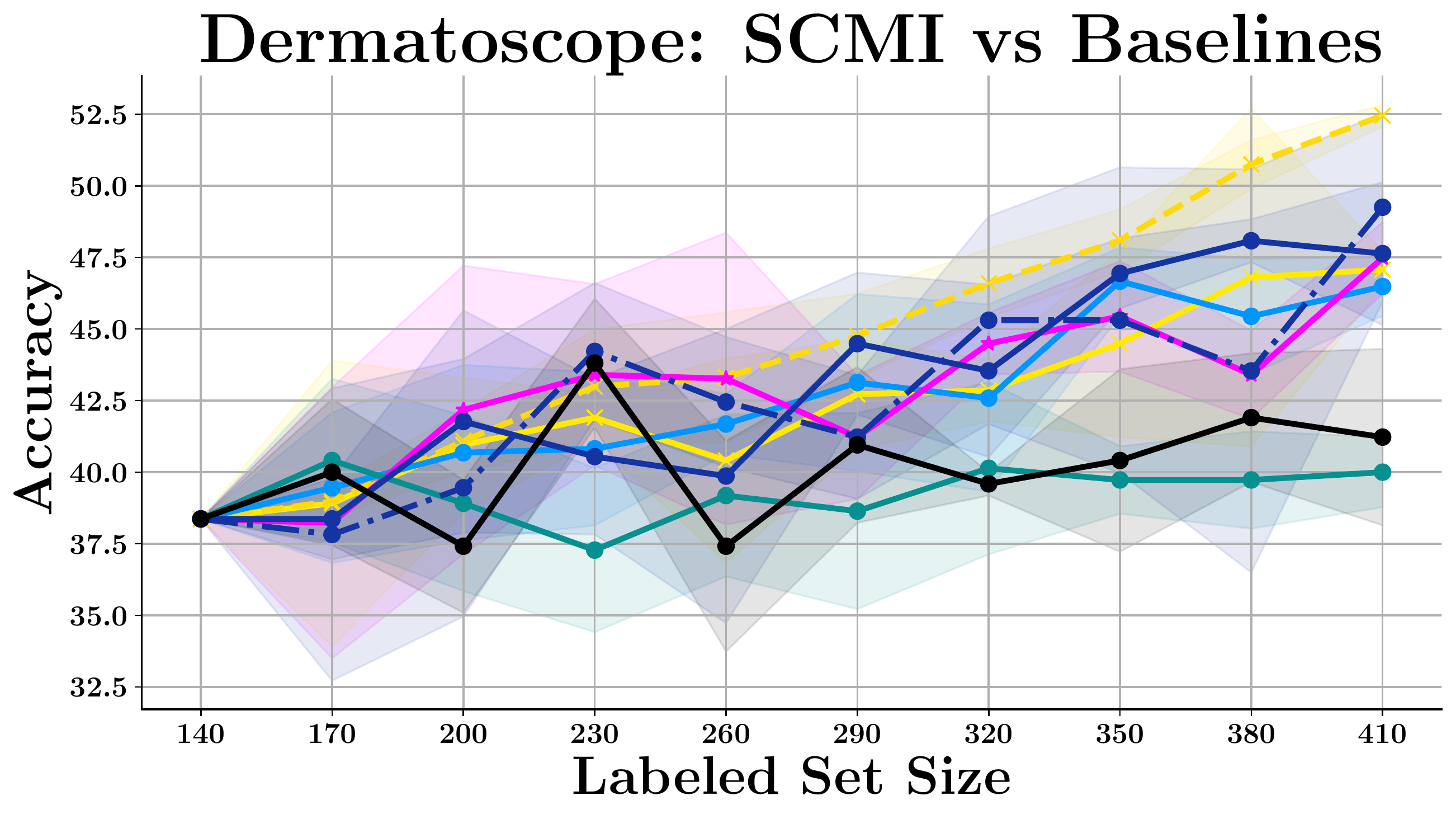}
\end{subfigure}
\begin{subfigure}[]{0.33\textwidth}
\includegraphics[width = \textwidth]{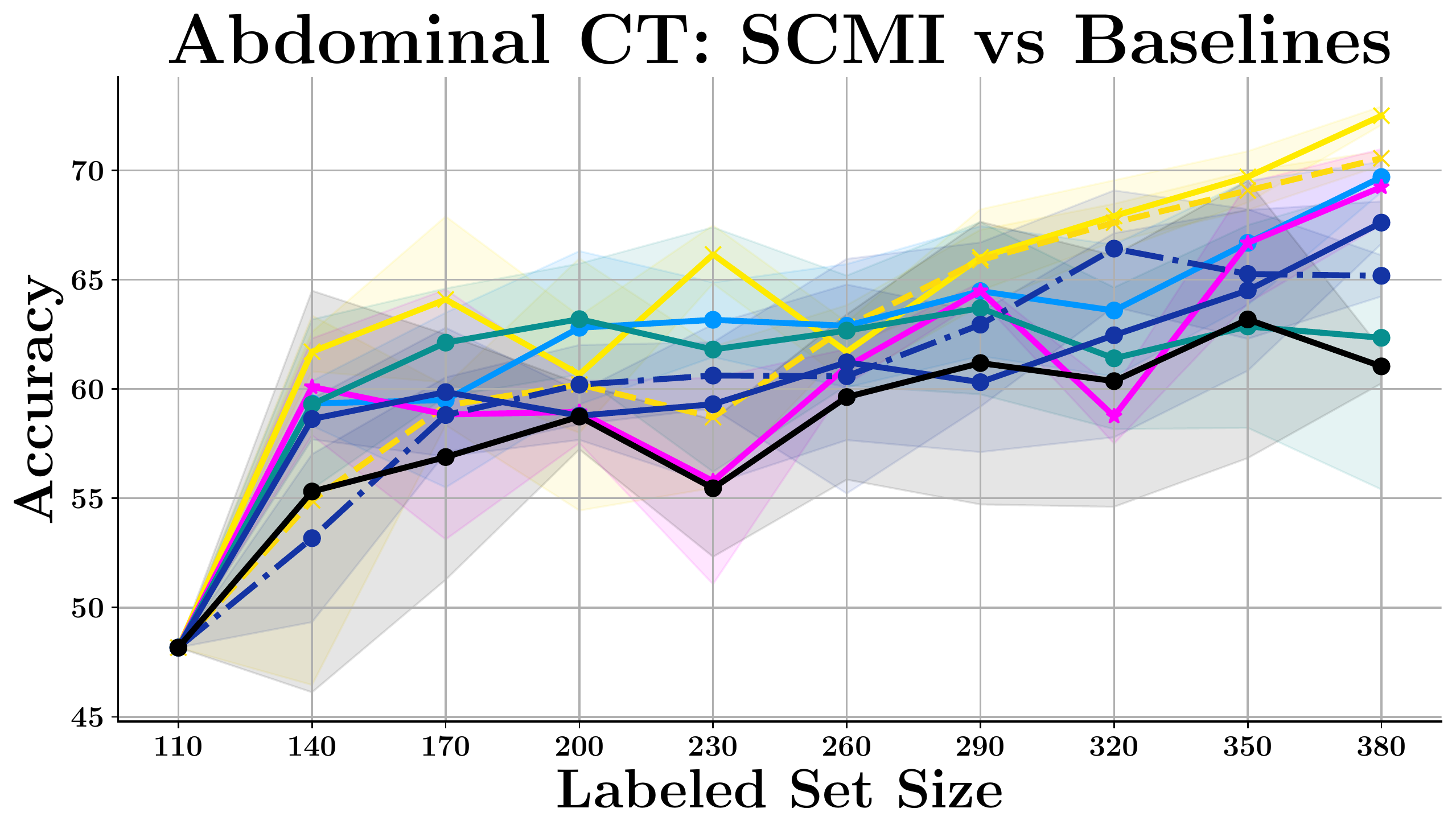}
\end{subfigure}
\begin{subfigure}[]{0.32\textwidth}
\includegraphics[width = \textwidth]{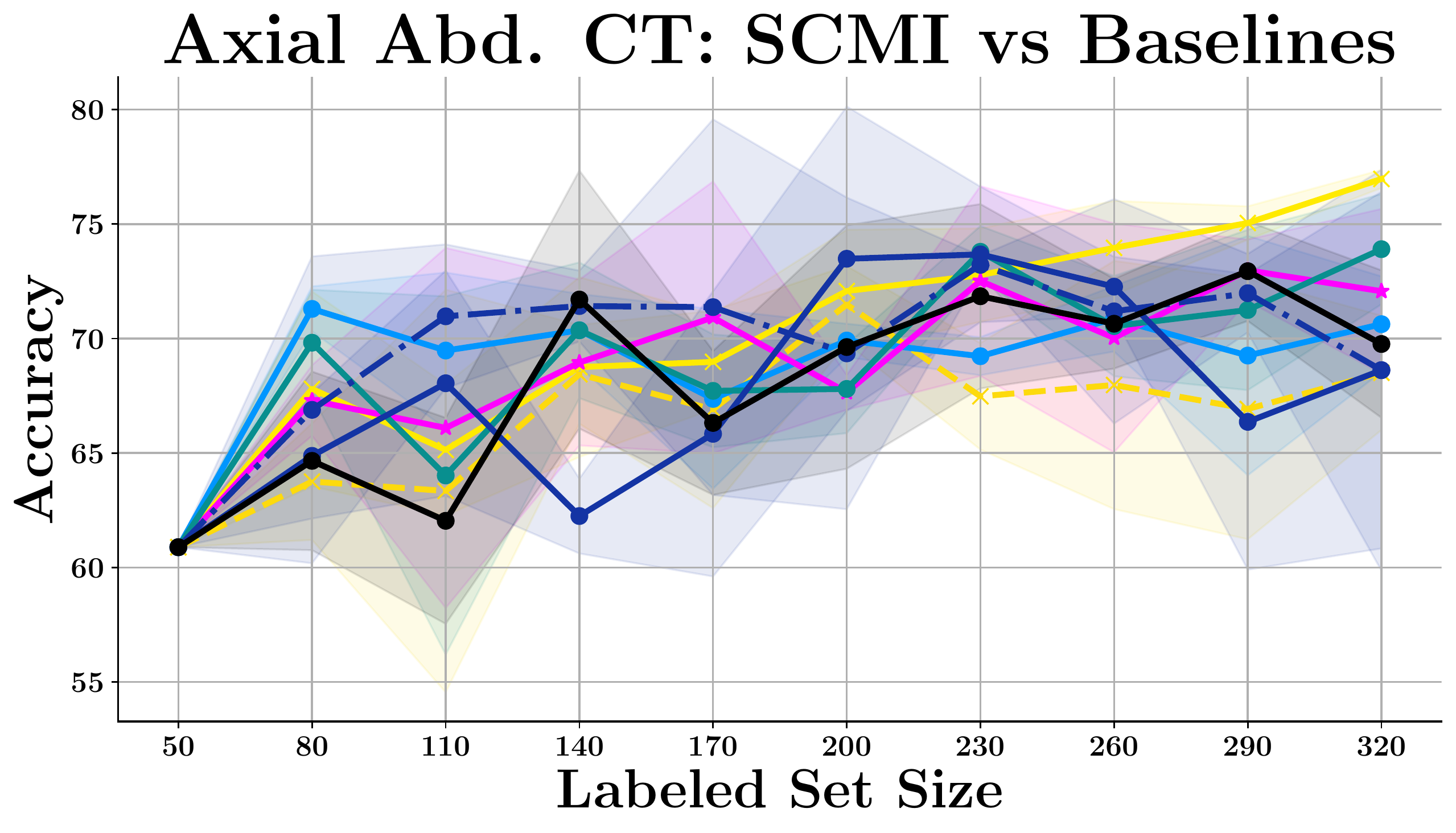}
\end{subfigure}
\begin{subfigure}[]{0.33\textwidth}
\includegraphics[width = \textwidth]{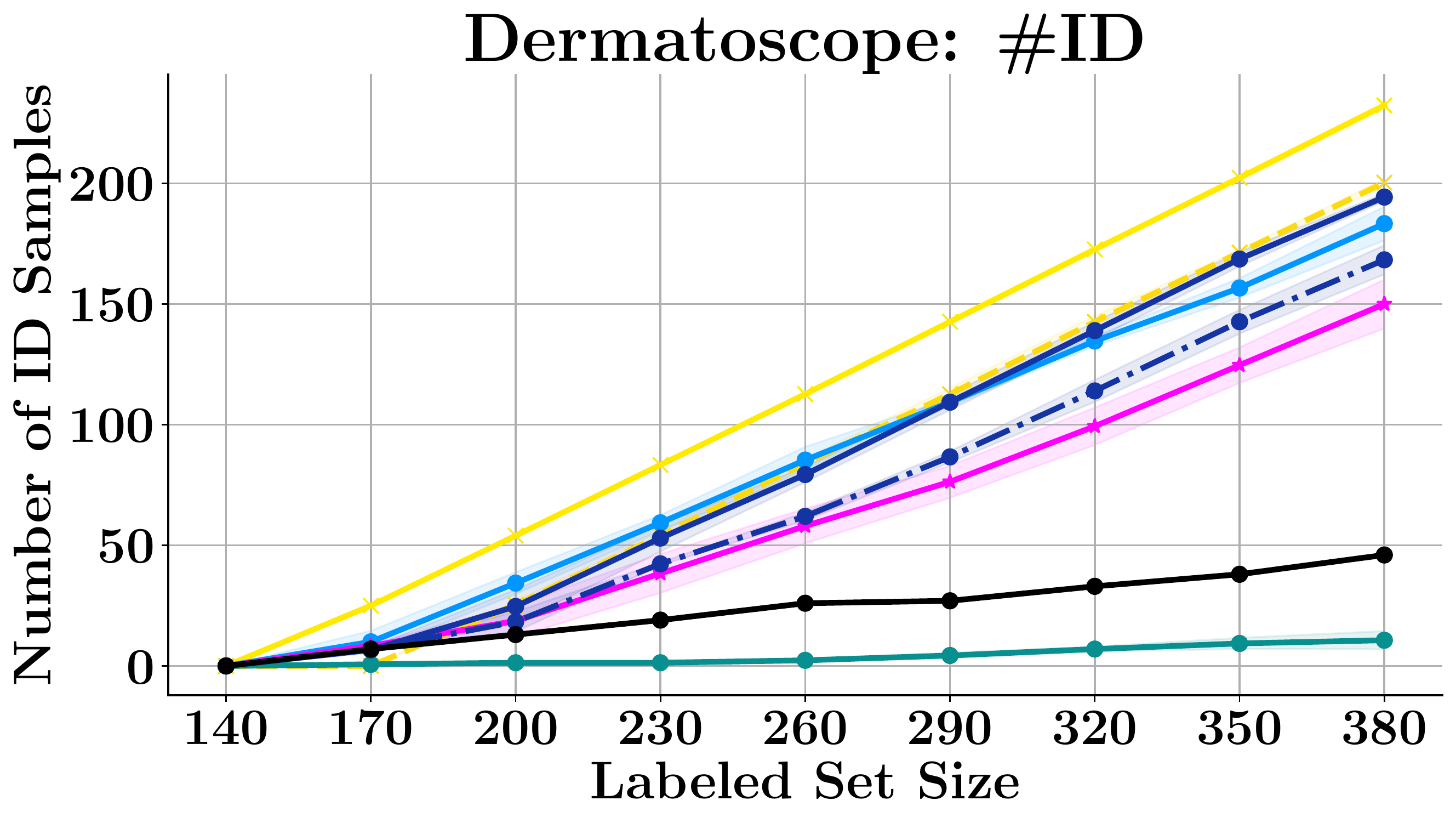}
\end{subfigure}
\begin{subfigure}[]{0.33\textwidth}
\includegraphics[width = \textwidth]{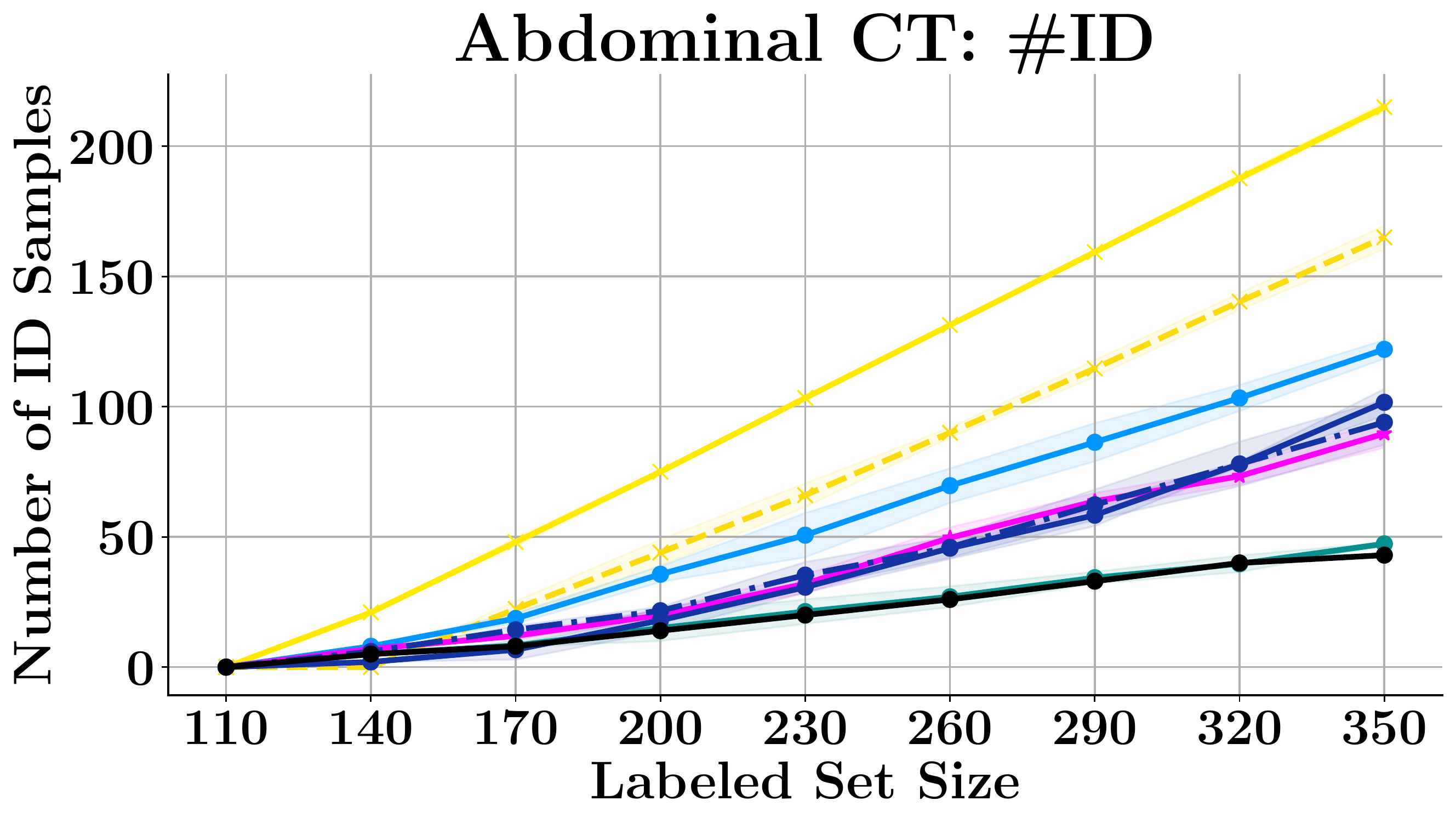}
\end{subfigure}
\begin{subfigure}[]{0.32\textwidth}
\includegraphics[width = \textwidth]{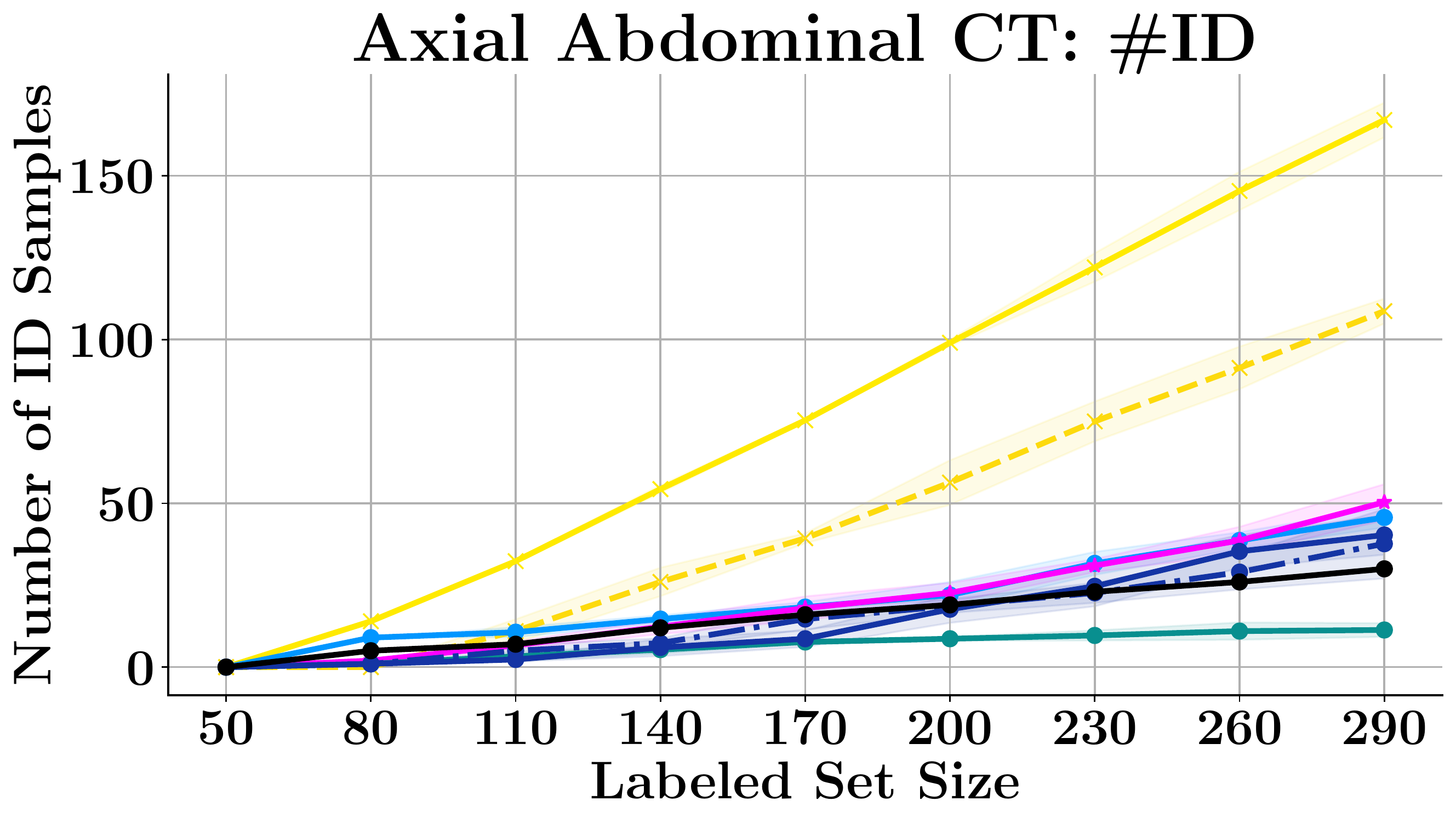}
\end{subfigure}
\caption{Active learning with medical OOD scenarios. \textbf{Top row:} \scmi\ vs Baselines. \textbf{Bottom row:} Number of ID points selected by each method. \textbf{First column - Scenario A:} We observe that facility location functions that balance representation and query-relevance are ideal for scenario A. Particularly, \textsc{Flcmi} consistently outperforms baselines by $\approx 5\% - 7\%$. \textbf{Second column - Scenario B:} The \scmi\ functions (\textsc{LogDetcmi, Flcmi}) outperform baselines by $\approx 4\% - 5\%$. \textbf{Third column - Scenario C:} We observe that \textsc{LogDetcmi} outperforms the baselines by $\approx 2\% - 4\%$. \textsc{LogDetcmi} selects the most number of ID points in all scenarios.}
\label{fig:res_ood_1}
\end{figure*}

\begin{figure*}[h]
\centering
\includegraphics[width = 12cm, height=0.7cm]{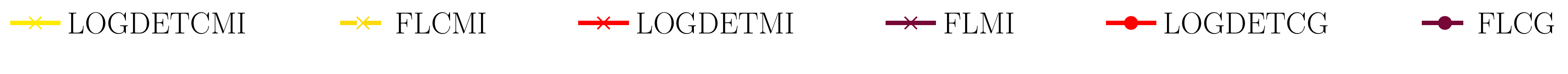}
\begin{subfigure}[]{0.33\textwidth}
\includegraphics[width = \textwidth]{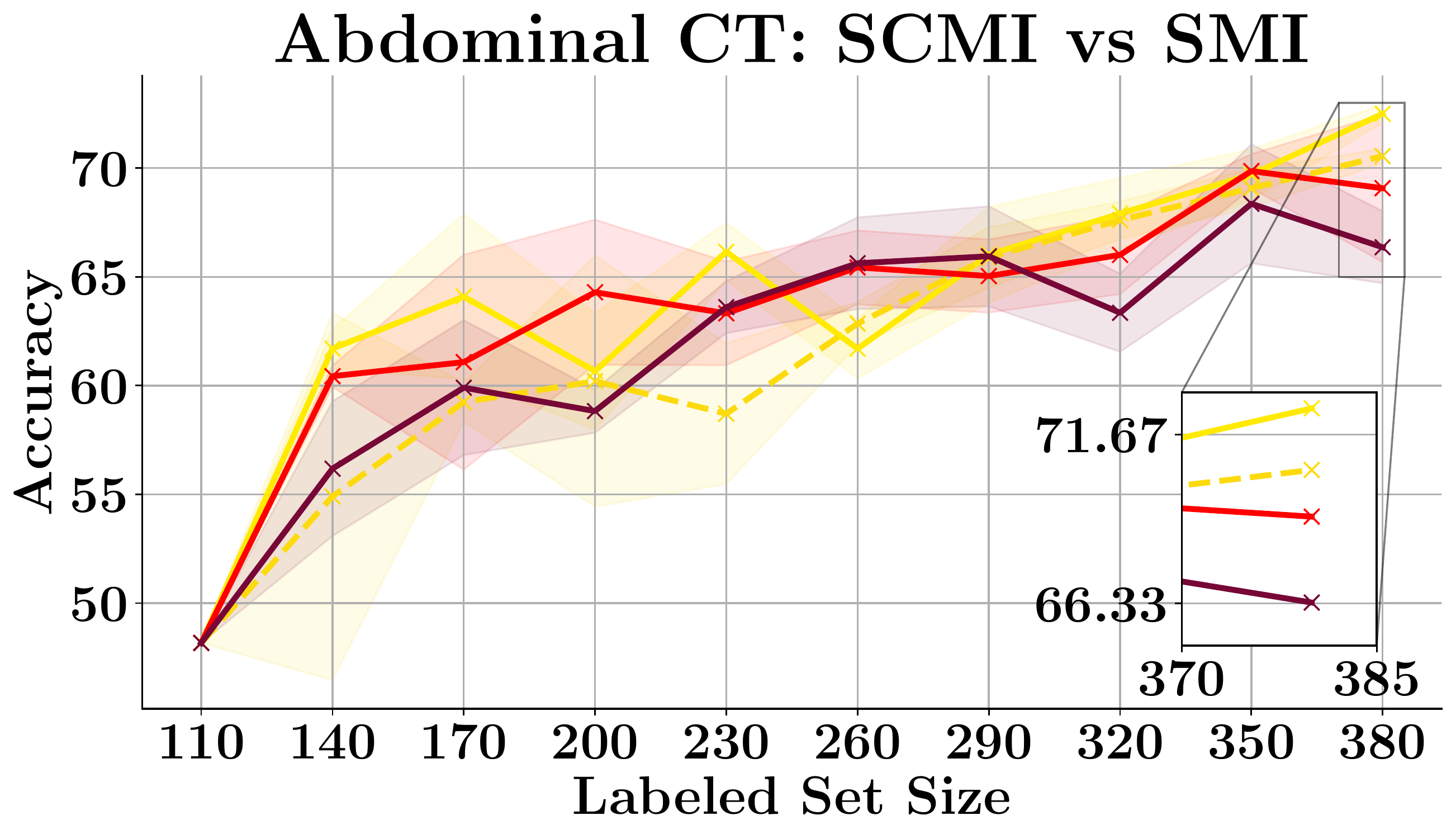}
\end{subfigure}
\begin{subfigure}[]{0.33\textwidth}
\includegraphics[width = \textwidth]{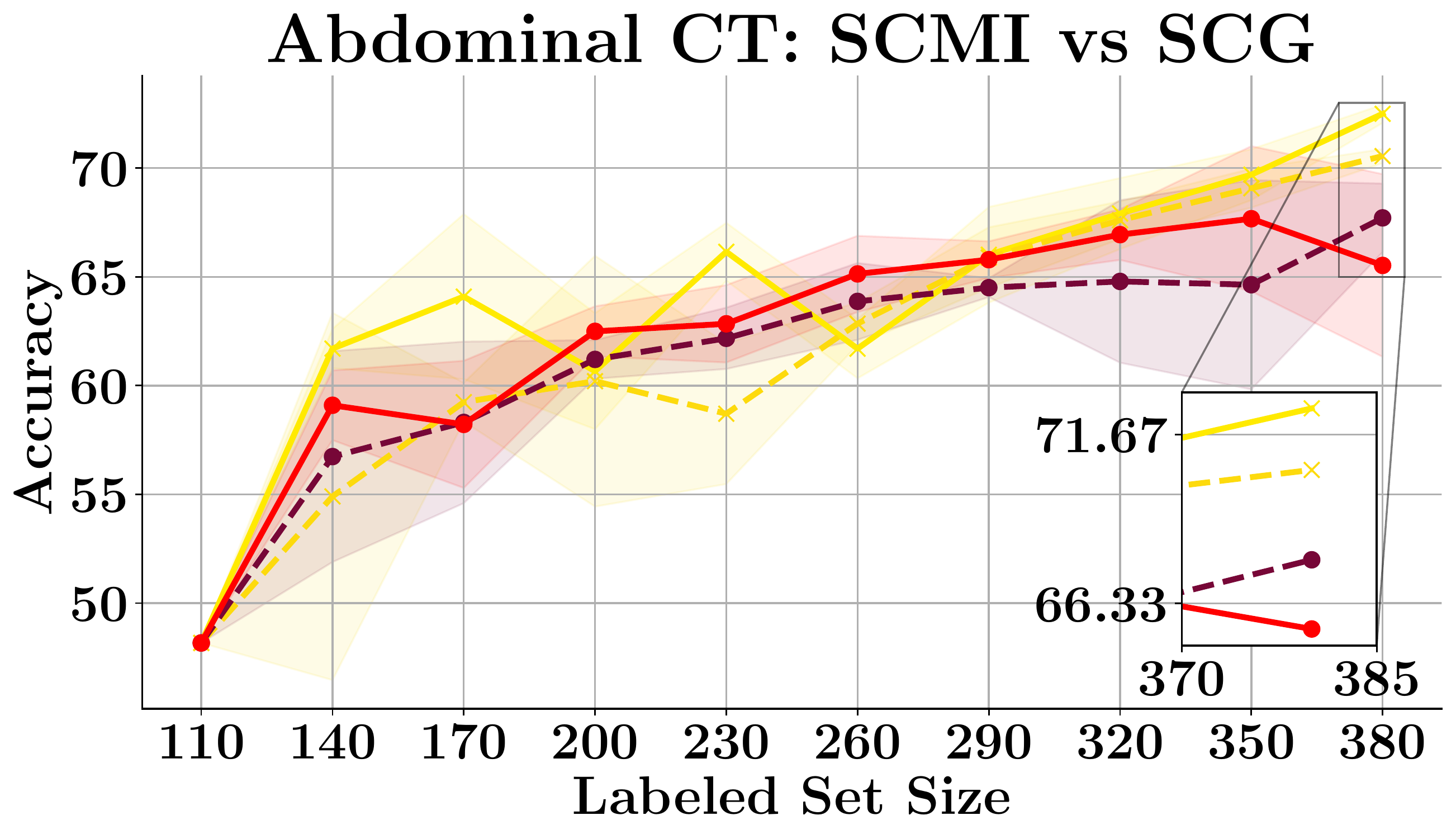}
\end{subfigure}
\begin{subfigure}[]{0.32\textwidth}
\includegraphics[width = \textwidth]{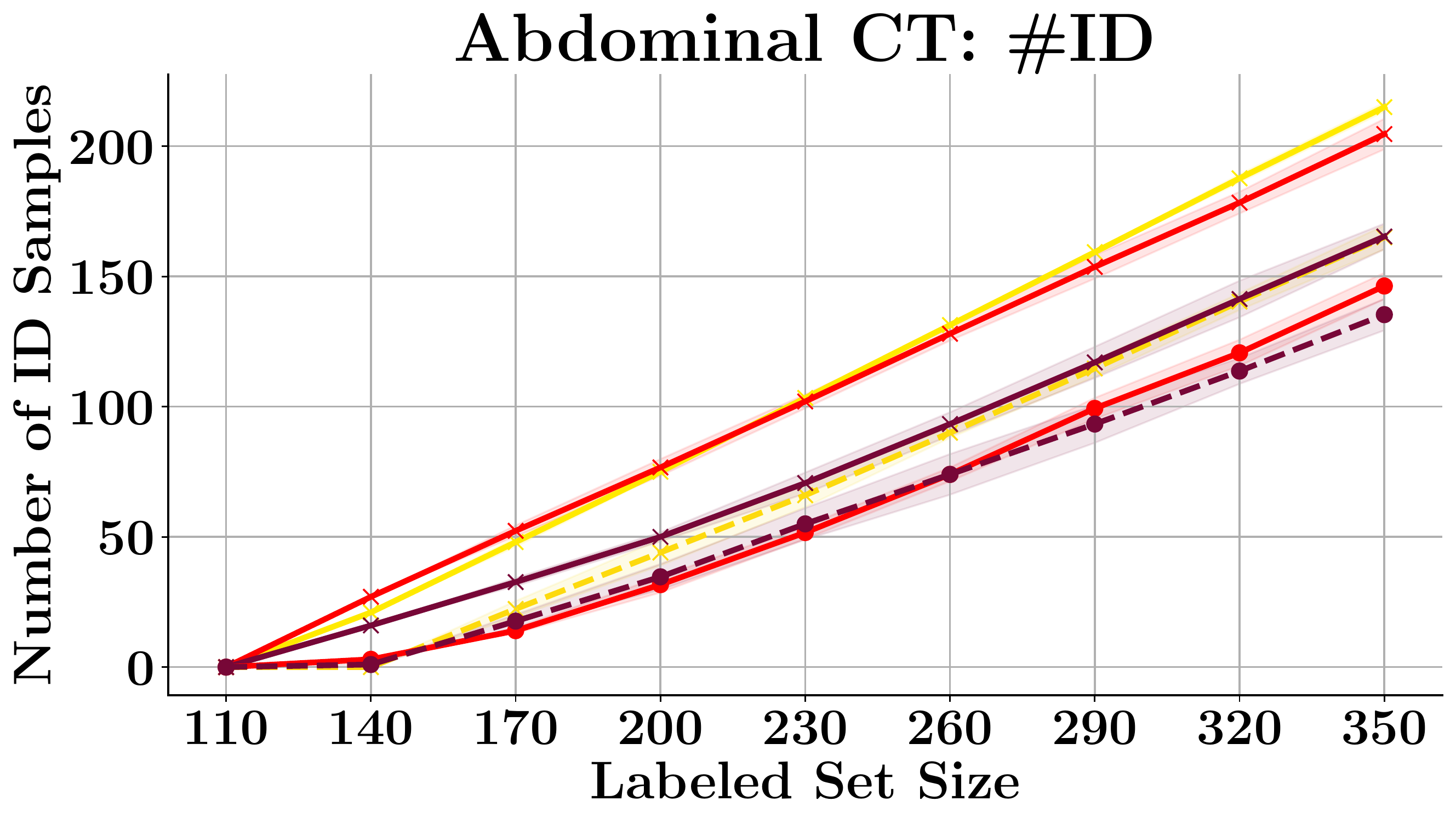}
\end{subfigure}
\caption{Ablation studies comparing the performance of \scmi\ functions with \smi\ functions (\textbf{left} plot) and \scg\ functions (\textbf{right} plot) for scenario B. We see that \scmi\ functions outperform their \smi\ and \scg\ counterparts, particularly in later rounds of AL. }
\label{fig:ablation_studies}
\end{figure*}

In this section, we evaluate the effectiveness of \model\ on three diverse medical imaging OOD data scenarios (A - C) with increasing levels of difficulty. We discuss these scenarios in detail in \secref{sec:ood_scenarios}. For evaluation, we compare the test accuracy and the number of in-distribution data points selected by the \scmi\ functions and existing state-of-the-art baselines in each round of active learning (see \figref{fig:res_ood_1}). We conduct ablation studies for each OOD data scenario to study the individual effect of only using a query set via the \smi\ functions and only using the private set via the \scg\ functions. We present the ablation study for one of the scenarios in (see \figref{fig:ablation_studies}) and defer the others to \Appref{app:ablation_studies}. In a nutshell, our experiments show that jointly modeling of similarity and dissimilarity using the \scmi\ functions not only outperforms the existing AL baselines but also the \smi\ and \scg\ functions across multiple OOD scenarios in medical data. In \Appref{app:pen_matrices}, we provide penalty matrices which show that \model\ statistically significantly outperforms the existing methods in all OOD scenarios. \looseness-1

\vspace{0.2cm}

\noindent \textbf{Baselines in all scenarios:} We compare the performance on \model\ against a variety of state-of-the-art uncertainty, diversity and targeted selection methods. The uncertainty based methods include \textsc{Entropy} and \textsc{Margin}. The diversity based methods include \textsc{Coreset} and \textsc{Badge}. For \textsc{Glister}, we maximize the log-likelihood with the set of ID points $\Ical$, for a fair comparison with the \scmi\ based acquisition functions. We discuss the details of all baselines in \secref{sec:related_work}. Lastly, we compare against random sampling (\textsc{Random}).

\vspace{0.2cm}

\noindent \textbf{Experimental setup: }We use the same training procedure and hyperparameters for all AL methods to ensure a fair comparison. For all experiments, we use a ResNet-18 \cite{he2016deep} model instantiated using (n+1) classes, where n is the number of ID classes and all other classes are grouped as a single OOD class. We train this model using an SGD optimizer with an initial learning rate of 0.001, the momentum of 0.9, and a weight decay of 5e-4. For each AL round, the weights are reinitialized using Xavier initialization and the model is trained till 99\% training accuracy. The learning rate is decayed using cosine annealing \cite{loshchilov2016sgdr} in every epoch. We run each experiment $5 \times$ on a V100 GPU and provide the error bars (std deviation). We discuss dataset splits for each of our experiments below and provide more details in \Appref{app:dataset_details}.

\vspace{-2ex}

\subsection{Scenario A - Unrelated Images} \label{sec:scenario_A}

\noindent \textbf{Dataset:}
In this scenario, we apply \model\ to avoid data points that are unrelated to the medical imaging domain. We use the Derma-MNIST (dermatoscopy of
pigmented skin lesions)~\cite{medmnistv2, kermany2018identifying} skin lesion image classification dataset as in-distribution (ID) data and CIFAR-10~\cite{krizhevsky2009learning} as OOD data. 
We create an initial labeled set $|\Lcal|=140$ using only ID data and an unlabeled set $\Ucal$ containing both ID data ($|\Ical_\Ucal|$=1061) and OOD data (|$\Ocal_\Ucal$|=5000) with AL batch size $B=30$.      

\vspace{0.2cm}

\noindent \textbf{Results:}
 We present results for the unrelated images OOD scenario in~\figref{fig:res_ood_1} (first column) and observe that the \textsc{Flcmi} consistently outperforms both uncertainty (\textsc{Entropy, Margin}) and diversity based (\textsc{Badge, Coreset}) methods by $\approx 5\% - 7\%$ on overall accuracy. Moreover, we observe that \textsc{Flcmi} outperforms \textsc{LogDetcmi} which suggests that using a submodular function like facility location that models \emph{representation} is useful for scenarios where the OOD data is obviously unrelated to the ID data. This also entails that a representative subset is imperative for obtaining a high accuracy on dermatoscopy modality datasets. \looseness-1

\vspace{-2ex}

\subsection{Scenario B - Incorrectly Acquired Images} \label{sec:scenario_B}

\noindent \textbf{Dataset:}
We apply \model\ to avoid CT scan images that are incorrectly prepared. Examples include images that are blurry, overexposed, underexposed or incorrectly cropped. We use OrganA-MNIST (Abdominal CT scans in an Axial plane)~\cite{medmnistv2, kermany2018identifying} organ image classification dataset as ID data. Following~\cite{cao2020benchmark}, we obtain OOD data by simulating different pre- and post-processing errors on CT scans such as inappropriate brightness, incorrect padding, cropping, and blurry images. We create the initial labeled set $|\Lcal|=110$ using only ID data and an unlabeled set $\Ucal$ containing both ID data ($|\Ical_\Ucal|$=1650) and OOD data (|$\Ocal_\Ucal$|=8000) with AL batch size $B=30$.

\vspace{0.2cm}
\noindent \textbf{Results:}
We present results for the incorrectly prepared medical images OOD scenario in \figref{fig:res_ood_1} (second column). We observe that the \scmi\ functions (\textsc{LogDetcmi, Flcmi}) outperform baselines by $\approx 4\% - 5\%$. The log determinant based functions that balance between \emph{diversity} and \emph{query-relevance} (\textsc{LogDetcmi, LogDetmi}) select the most number of ID data points and perform well in this scenario.

\vspace{0.2cm}

\noindent \textbf{Ablation study:} Interestingly, the conditional gain functions (\textsc{Flcg, LogDetcg}) do not select as many ID points but still perform at par with the \smi\ functions (see \figref{fig:ablation_studies}). This suggests the need for conditioning in difficult OOD scenarios where the ID and OOD points have a high semantic similarity. Hence, jointly maximizing the conditional gain and mutual information is imperative, as done in the \scmi\ functions.


\vspace{-2ex}

\subsection{Scenario C - Mixed View Images} \label{sec:scenario_C}

\noindent \textbf{Dataset:}
We apply \model\ to avoid Abdominal CT scan images that are captured from a different view of the anatomy. We use OrganA-MNIST (axial plane)~\cite{medmnistv2, kermany2018identifying} organ image classification dataset as ID data, and a combination of OrganC-MNIST (coronal plane)~\cite{medmnistv2, kermany2018identifying} and OrganS-MNIST (sagittal plane)~\cite{medmnistv2, kermany2018identifying} as OOD data. We create the initial labeled set $|\Lcal|=50$ using only ID data an unlabeled set $\Ucal$ containing both ID data ($|\Ical_\Ucal|$=750) and OOD data (|$\Ocal_\Ucal$|=8000) with AL batch size $B$=30.

\vspace{0.2cm}

\noindent \textbf{Results:} We present results for the mixed view medical images OOD scenario in \figref{fig:res_ood_1}(third column). We observe that \textsc{LogDetcmi} outperforms the baselines by $\approx 2\% - 4\%$. We observe from scenarios B and C that the log determinant functions select significantly more ID data points from the unlabeled set and outperform other methods when the modality is CT. This entails that selecting a diverse subset is one of the key factors for CT modality data.


\vspace{-2ex}

\section{Conclusion} \label{sec:conclusion}
We demonstrate the effectiveness of \model\ across a diverse set of out-of-distribution (OOD) scenarios in medical data. We observe that \scmi\ functions outperform other baselines along with \smi\ and \scg\ functions. Which submodular function works best depends on the modality of medical data and the type of OOD scenario. Importantly, we note that jointly maximizing both components, mutual information and conditional gain, works the best for scenarios with OOD data. Lastly, as expected, we observe a drop in accuracy gain as the difficulty of OOD scenarios increases. \looseness-1

\bibliographystyle{splncs04}
\bibliography{main}

\newpage

\appendix

\setcounter{page}{1}

\section*{Supplementary Material for \model: Avoiding Out-of-distribution Data using Submodular Information Measures} 

\section{Summary of Notations}\label{app:notation-summary}

 \begin{table*}[!h]
 \centering
 \begin{tabular}{|l|l|p{0.5\textwidth}|} 
 \toprule
 \hline 
 \multicolumn{1}{|l|}{Topic} & Notation & Explanation \\ \hline
 \toprule \hline
 & ID & In-Distribution \\
 & OOD & Out-Of-Distribution\\
 &  $\Ucal$ & Unlabeled set of $|\Ucal|$ instances\\ 
 \multicolumn{1}{|p{0.20\textwidth}|}{ \model\ (\secref{sec:our_method})} 
 & $\Acal$ & A subset of $\Ucal$\\ 
 & $S_{ij}$ & Similarity between any two data points $i$ and $j$\\
 & $f$ & A submodular function\\
 & $\Lcal$ & Labeled set of data points\\
 & $\Qcal$ & Query set\\
 & $\Pcal$ & Private set\\
 & $\Mcal$ & Deep model\\
 & $B$ & Active learning selection budget\\
 & $\Hcal$ & Loss function used to train model $\Mcal$\\
 & $\Xcal$ & Pairwise similarity matrix computed using gradients\\
 & $\Ical$ & Set of in-distribution data points $\Ical \subseteq \Lcal$\\
 & $\Ocal$ & Set of out-of-distribution data points $\Ocal \subseteq \Lcal$\\
 & $\Gcal_\Acal$ & Gradients of some subset $\Acal$\\
 \hline
 \bottomrule
 \end{tabular}
 \caption{Summary of notations used throughout this paper}
 \label{tab:main-notations}
 \end{table*}

\section{Details of Datasets used}

\subsection{Derma-MNIST~\cite{medmnistv2, kermany2018identifying}}

\begin{itemize}
    \item The Derma-MNIST is a large collection of multi-source dermatoscopic images of common pigmented skin lesions.
    \item The dataset consists of 10,015 dermatoscopic images of size 3$\times$28$\times$28 categorized as 7 classes 
    \item Classes represent various types of skin diseases namely - Melanocytic nevi, Melanoma, Benign keratosis-like lesions, Basal cell carcinoma, Actinic keratoses, Vascular lesions, Dermatofibroma) 
    
\end{itemize}

\subsection{CIFAR-10~\cite{krizhevsky2009learning}}

\begin{itemize}
    \item CIFAR-10  is a collection of images of objects namely - airplanes, cars, birds, cats, deer, dogs, frogs, horses, ships, and trucks.
    \item The dataset consists of 60,000 images of size 3$\times$32$\times$32 categorized into 10 classes, with 6,000 images per class.
\end{itemize}

\subsection{OrganMNIST~\cite{medmnistv2, kermany2018identifying}}

\begin{itemize}
    \item The Organ\{A,C,S\}MNIST is based on 3D computed tomography (CT) images from Liver Tumor Segmentation Benchmark
(LiTS).  Hounsfield-Unit (HU) of the 3D images
are transformed into grey scale with a abdominal window; then 2D images are cropped from the center slices of the 3D bounding boxes in respective views (planes).
    \item The only difference among Organ\{A,C,S\}MNIST is the views - Axial,Coronal and Sagittal. 
    \item The dataset consists of images of size 1$\times$28$\times$28 categorized as 11 classes 
    \item Classes represent various organs namely - heart, left lung, right lung, liver, spleen, pancrea, left kidney, right kidney, bladder, left femoral head and right femoral head.
    
\end{itemize}

\label{app:dataset_details}

\section{Scalability of \model} \label{app:scalability}

Below, we provide a detailed analysis of the complexity of creating and optimizing the different SIM functions. Denote $|\Xcal|$ as the size of set $\Xcal$. Also, let $|\Ucal| = n$ (the ground set size, which is the size of the unlabeled set in this case)

\begin{itemize}
    \item \textbf{Facility Location: } For \textsc{Flvmi}, the complexity of creating the kernel matrix is $O(n^2)$. The complexity of optimizing it is $\tilde{O}(n^2)$ (using memoization~\cite{iyer2019memoization})\footnote{$\tilde{O}$: Ignoring log-factors} if we use the stochastic greedy algorithm~\cite{mirzasoleiman2015lazier} and $O(n^2k)$ with the naive greedy algorithm. The overall complexity is $\tilde{O}(n^2)$. For FLCMI, the complexity of computing the kernel matrix is $O([n + |\Qcal| + |\Pcal|]^2)$, and the complexity of optimization is $\tilde{O}(n^2)$.
    \item \textbf{Log-Determinant: } We start with LogDetMI. The complexity of the kernel matrix computation (and storage) is $O(n^2)$. The complexity of optimizing the LogDet function using the stochastic greedy algorithm is $\tilde{O}(B^2 n)$, so the overall complexity is $\tilde{O}(n^2 + B^2n)$. For LogDetCG, the complexity of computing the matrix is $O([n + |\Pcal|]^2$, and the complexity of optimization is $\tilde{O}([B + |\Pcal|]^2 n)$. For the LogDetCMI function, the complexity of computing the matrix is $O([n + |\Pcal| + |\Qcal|]^2$, and the complexity of optimization is $\tilde{O}([B + |\Pcal| + |\Qcal|]^2 n)$. 
\end{itemize}

We end with a few comments. First, most of the complexity analysis above is with the stochastic greedy algorithm~\cite{mirzasoleiman2015lazier}. If we use the naive or lazy greedy algorithm, the worst-case complexity is a factor $B$ larger. Secondly, we ignore log-factors in the complexity of stochastic greedy since the complexity is actually $O(n\log 1/\epsilon)$, which achieves an $1 - 1/e - \epsilon$ approximation. Finally, the complexity of optimizing and constructing the FL and LogDet functions can be obtained from the CG versions by setting $\Pcal = \emptyset$.

\section{Additional Results}  \label{app:add_res}

\subsection{\model\ for Domain generalization} \label{app:scenario_D}

\begin{figure*}
\hspace*{-2cm}                                                                                                 
\includegraphics[width = 13.8cm, height=8cm]{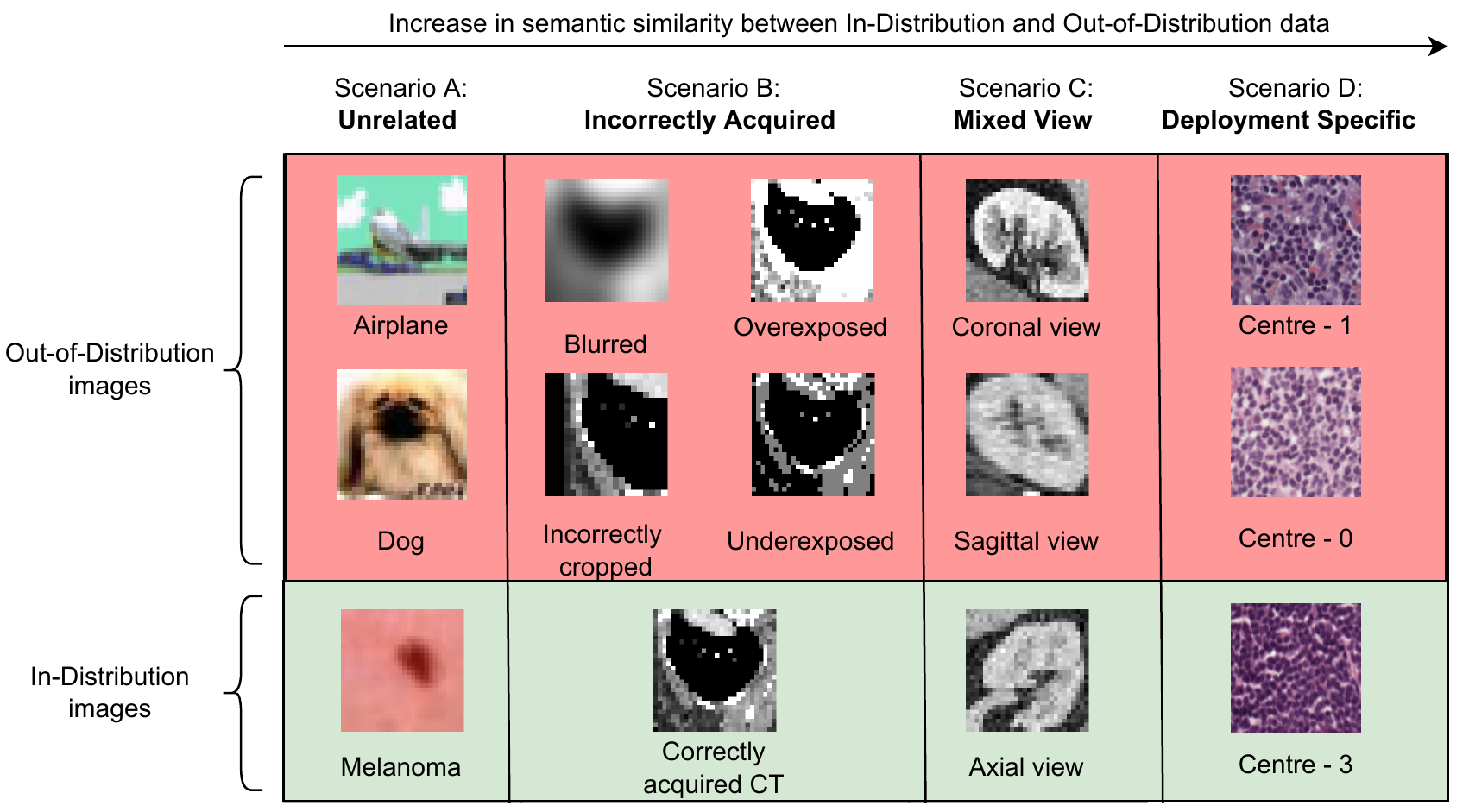}

\caption{The out-of-distribution (OOD) images in first three scenarios are contrasted with the in-distribution (ID) images.       
A: Inputs that are unrelated to the task.
B: Inputs which are incorrectly acquired.
C: Inputs that belong to a different view of anatomy.
D: Inputs collected at a different centre. Note that these scenarios become \emph{increasingly} difficult as we go from A $\rightarrow$ D since the semantic similarity between OOD and ID increases.}
\label{fig:scenarios_4}
\end{figure*}

\noindent \textbf{Scenario D - Domain Generalization:} Avoid images that are \emph{not} aligned with a deployment specific target. For instance, images acquired in a hospital belonging to a different demography than the deployment. In the Scenario D of \figref{fig:scenarios_4}, data points from Centre- 0 and Centre- 1 are OOD when the machine learning model is to be deployed at Centre- 3. As pointed out in \cite{miller2021accuracy}, this scenario is challenging since the semantics of the data may vary only in some aspects across different demographics. However, these demography based aspects may be critical for training a better model. 

\noindent \textbf{Dataset:}
For this scenario we apply our framework to acquire images from any hospitals(target hospital(s)) where the final model will be deployed. We use patch-based variant of Camelyon17 dataset \cite{wilds2021} consisting of metasized breast cancer tissues from five different hospitals\cite{bandi2018detection}. The initial labeled set $\Lcal$ (seed set) in AL consists of OUT data from Centres 0,3,4 as defined in \cite{wilds2021}. Centre 1 is considered as IN data. Assumption here is that a model initially trained with data from 3 different hospitals, is to be fine-tuned for deployment in Centre 1 . In the original study in \cite{wilds2021} the dataset was considered for domain generalization setting. In our case, we seek to maximize the gain from both IN and OOD data, as they are semantically similar such that the model performs well for the target hospital. The data selection is guided by a query set $\Qcal$ representing the distribution of the target hospital. $\Qcal$ consists of two data points, one for each of the classification(tumor/non-tumor), from each of the ten whole-slide images(WSI) of the target hospital .

\begin{figure*}[h]
\centering
\includegraphics[width = 12cm, height=0.7cm]{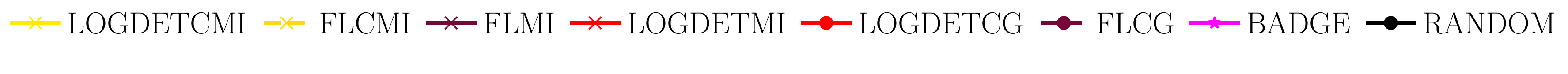}
\begin{subfigure}[]{0.48\textwidth}
\includegraphics[width = \textwidth]{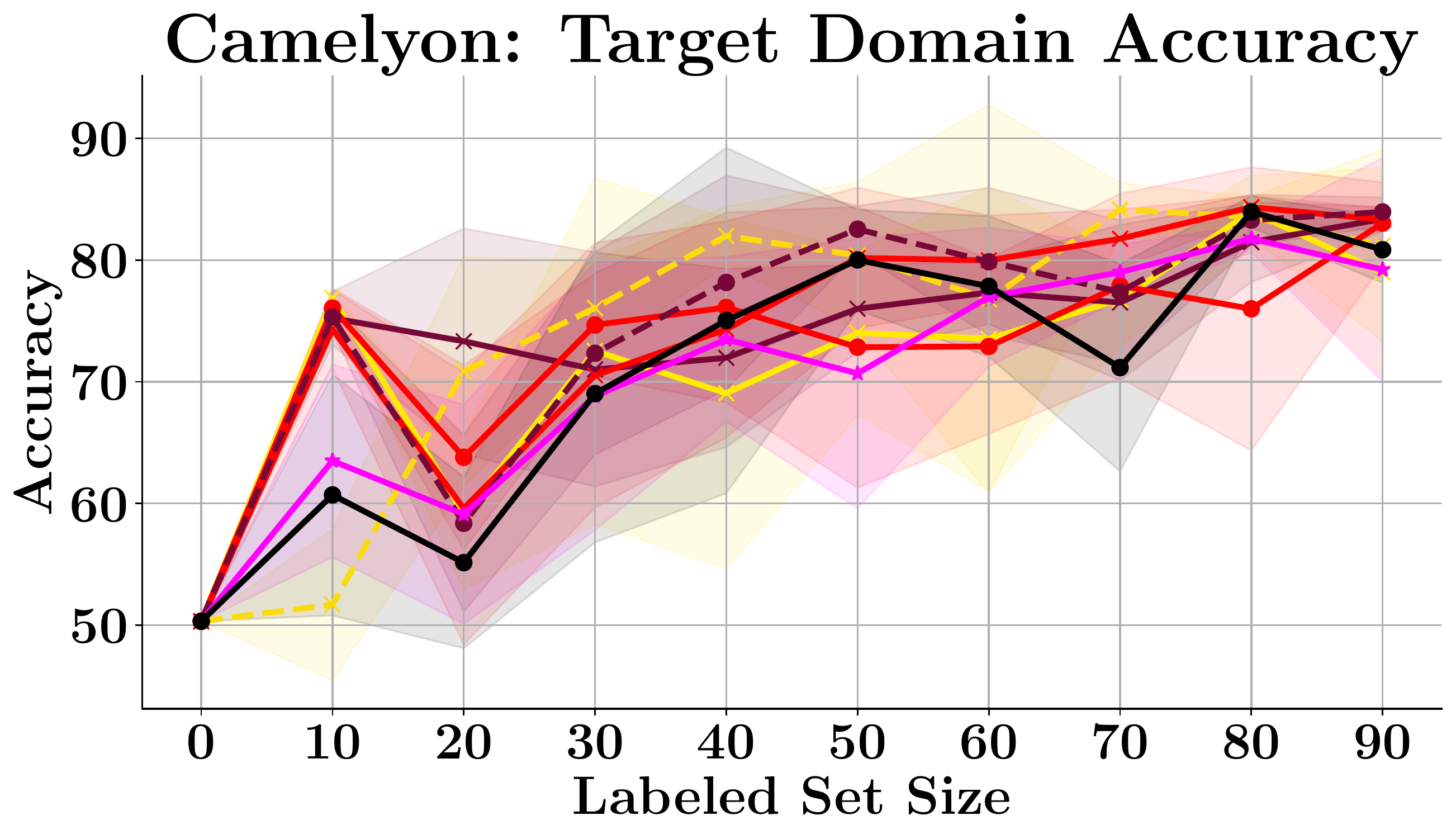}
\end{subfigure}
\begin{subfigure}[]{0.48\textwidth}
\includegraphics[width = \textwidth]{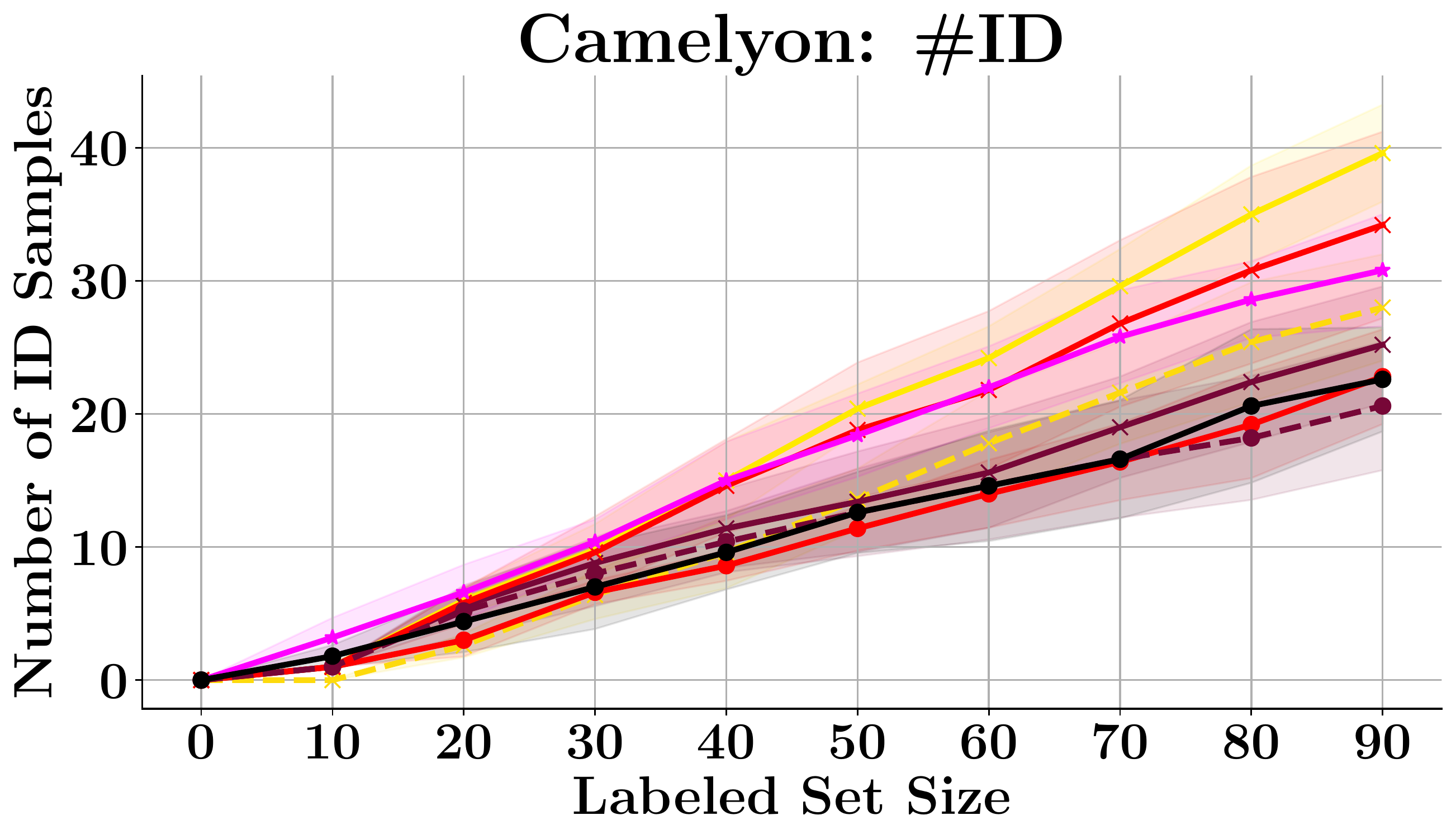}
\end{subfigure}
\caption{The \scmi\ functions, in particular \textsc{LogDetcmi} outperforms baselines for selecting datapoints from target centre. Yet, as observed \cite{miller2021accuracy} the performance is significant variation on the performance even over multiple runs }
\label{fig:oodresults}
\end{figure*}



\noindent \textbf{Results:}
We present results for the deployment specific medical images OOD scenario in \ref{fig:oodresults}. We observe that the \scmi\ functions outperform other baselines in selecting datapoints from target centre almost by twice the numbers. Yet the performance of model is highly fluctuating as seen in the plot. This could be due to the high degree of co-relation within a hospital or slides which result in training and evaluation performance variability. The general approach to overcome this is to use model specific settings during the training along with our DIAGONSE framework whenever we want the models to be deployed to a target hospital.

\subsection{Ablation studies} \label{app:ablation_studies}
\begin{figure*}[h]
\centering
\includegraphics[width = 12cm, height=0.7cm]{new_plots/cmi_mi_cg_legend.pdf}
\begin{subfigure}[]{0.33\textwidth}
\includegraphics[width = \textwidth]{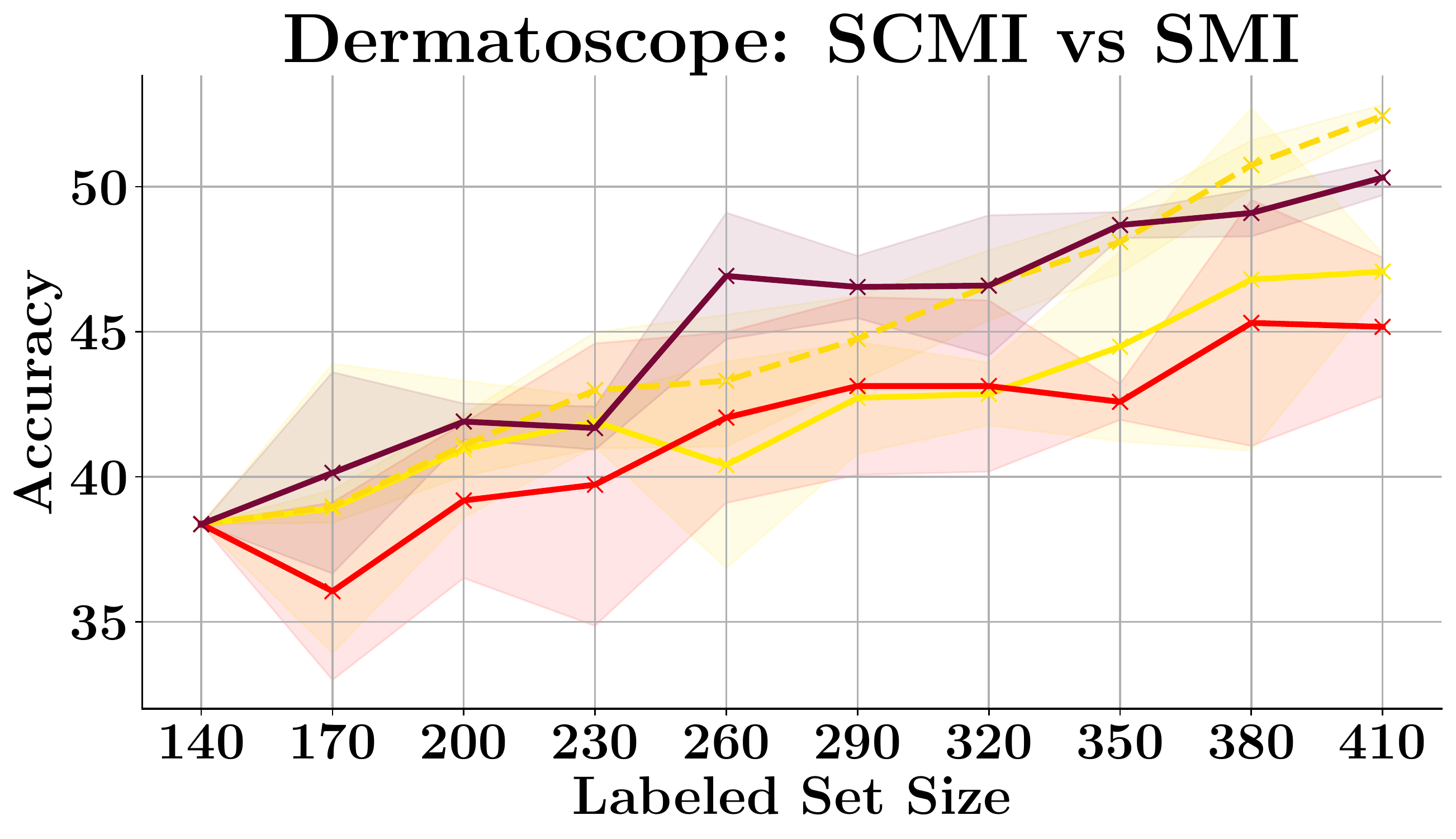}
\end{subfigure}
\begin{subfigure}[]{0.33\textwidth}
\includegraphics[width = \textwidth]{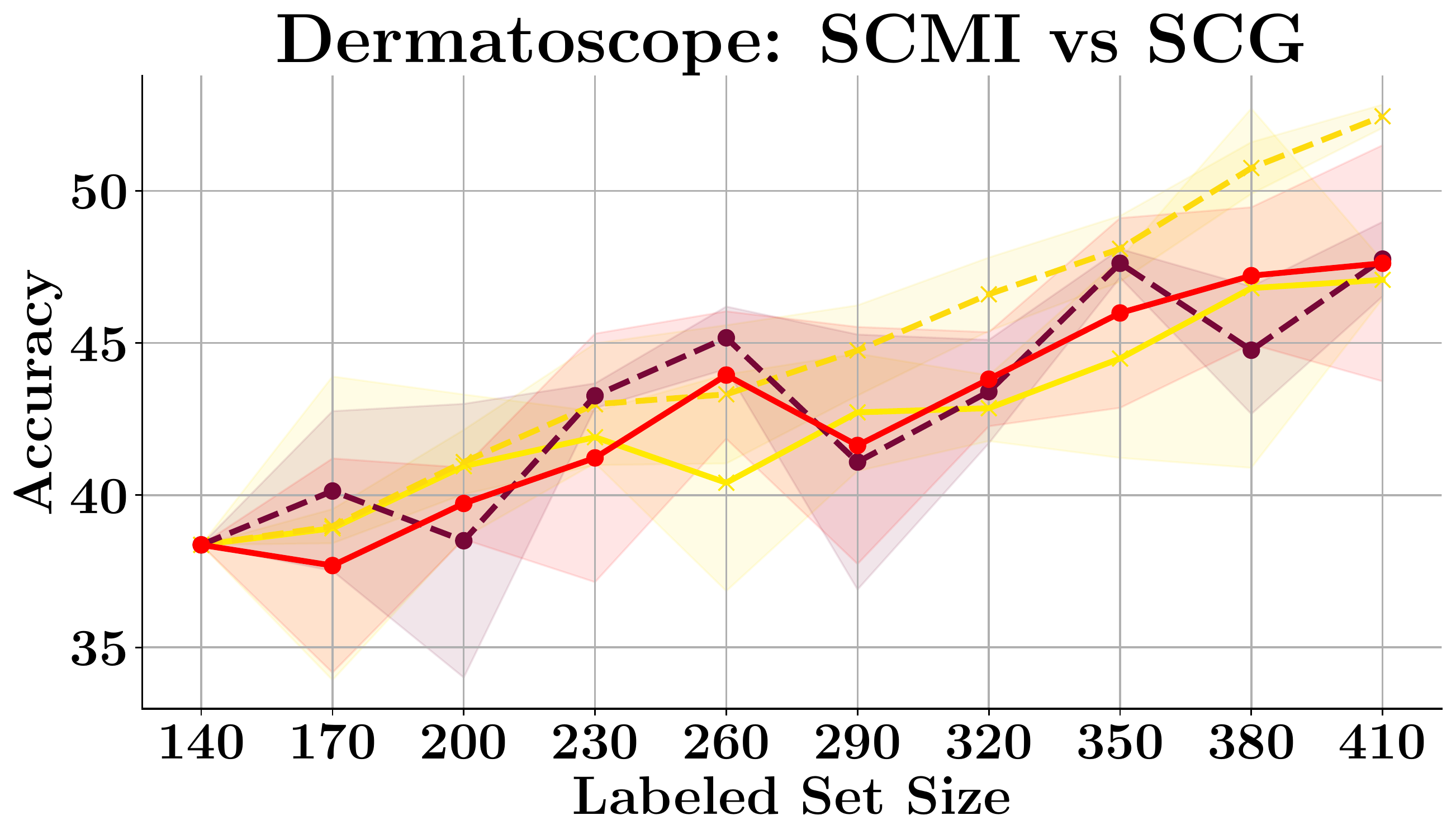}
\end{subfigure}
\begin{subfigure}[]{0.32\textwidth}
\includegraphics[width = \textwidth]{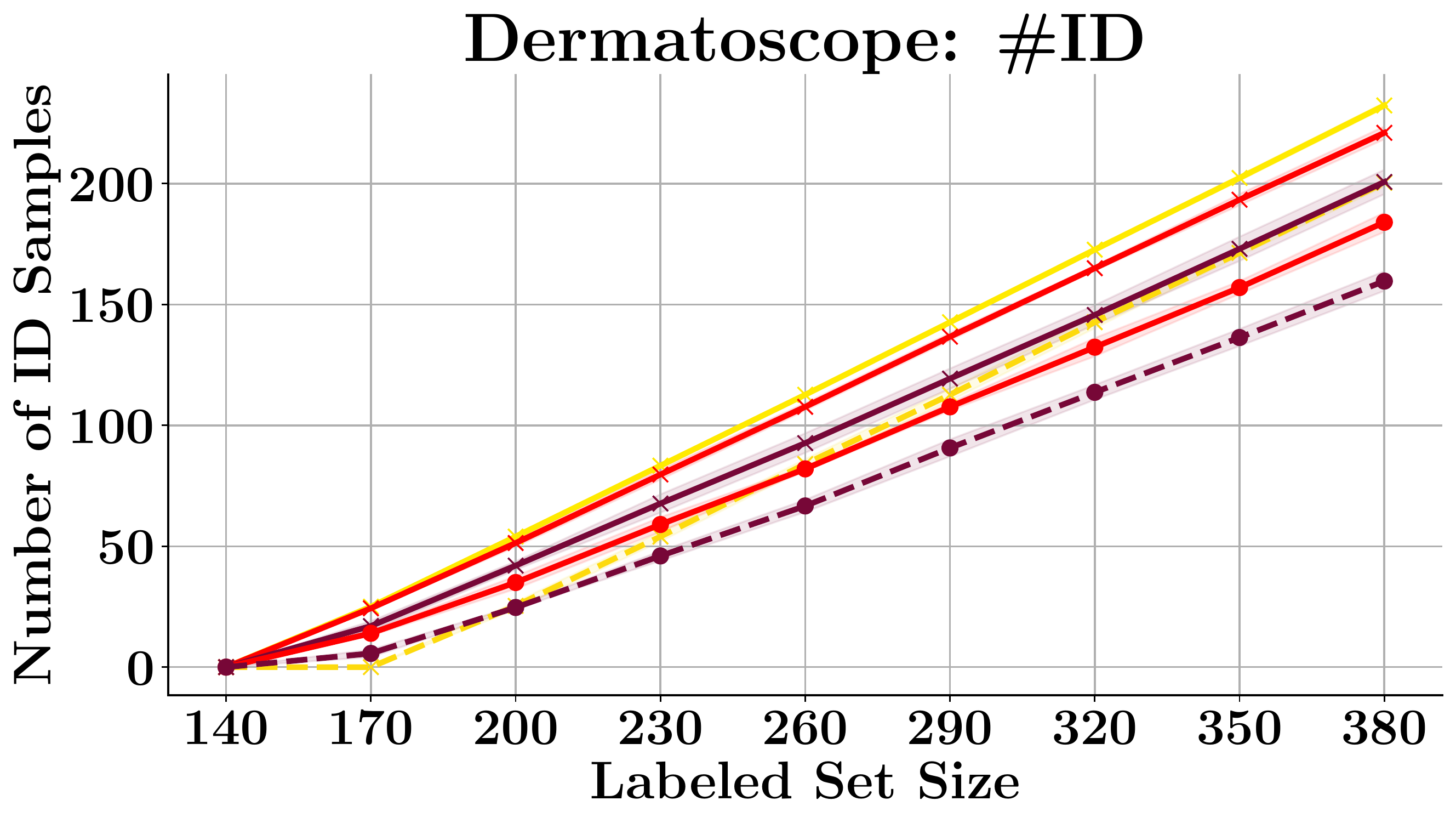}
\end{subfigure}
\begin{subfigure}[]{0.33\textwidth}
\includegraphics[width = \textwidth]{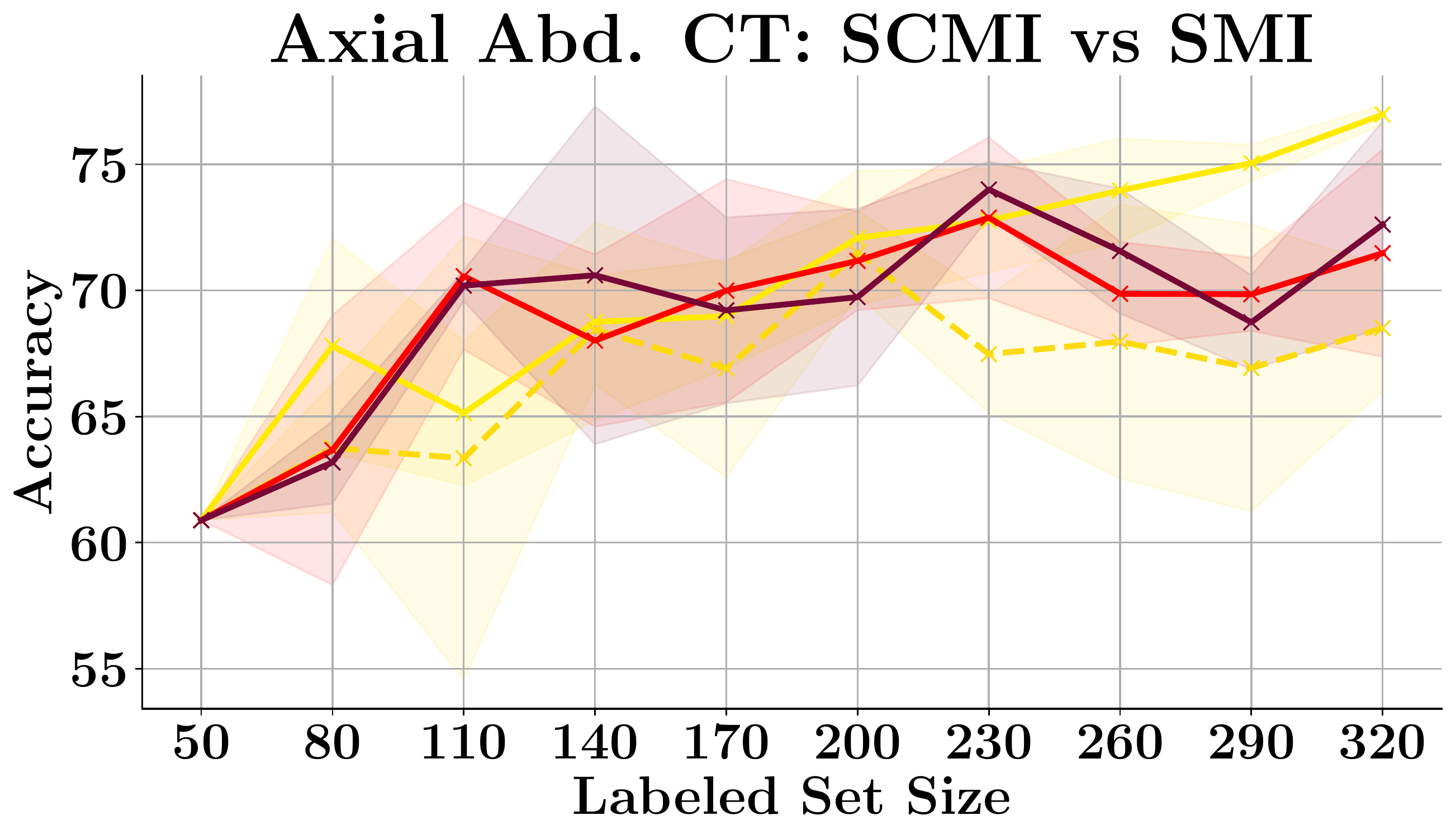}
\end{subfigure}
\begin{subfigure}[]{0.33\textwidth}
\includegraphics[width = \textwidth]{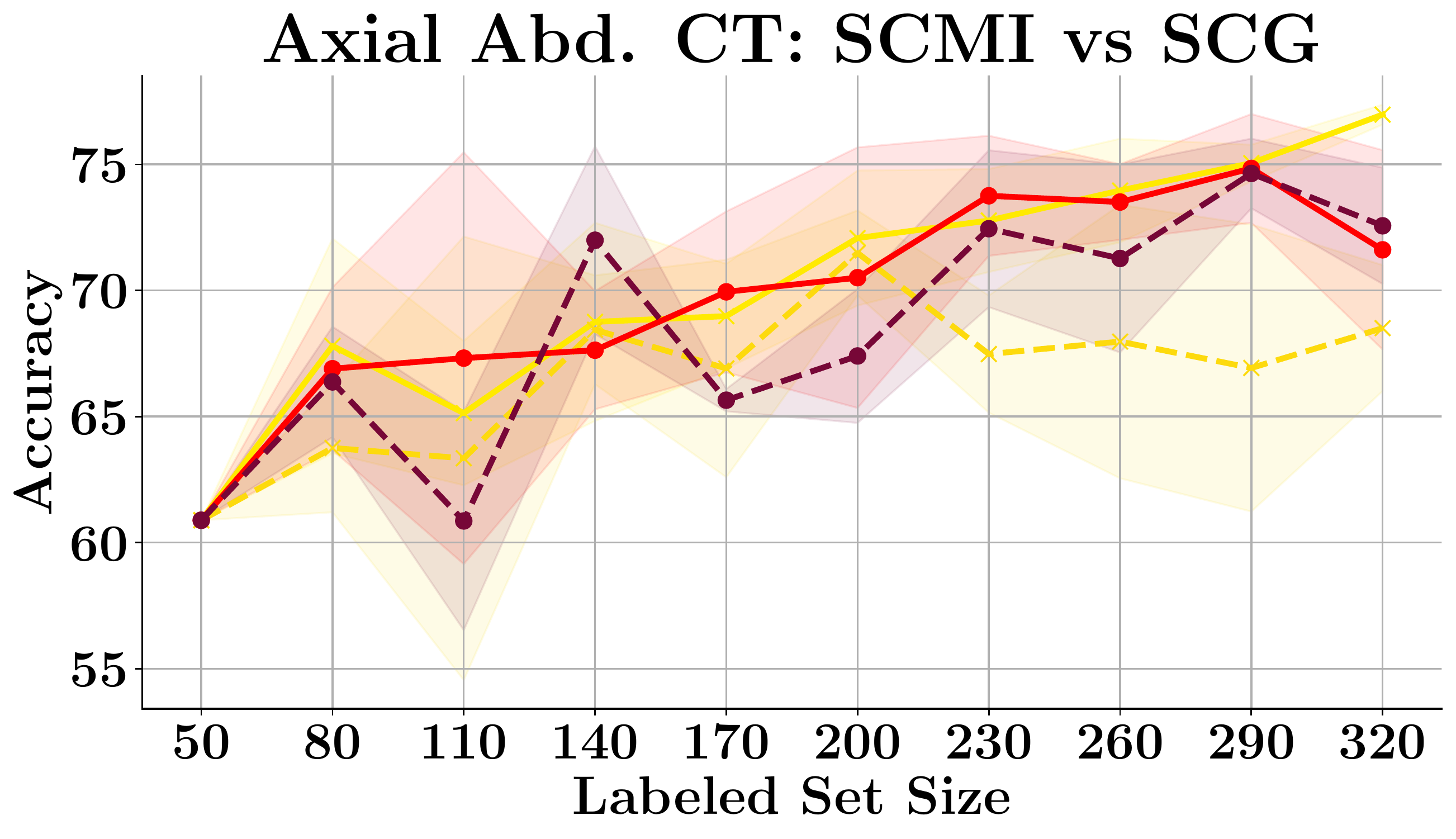}
\end{subfigure}
\begin{subfigure}[]{0.32\textwidth}
\includegraphics[width = \textwidth]{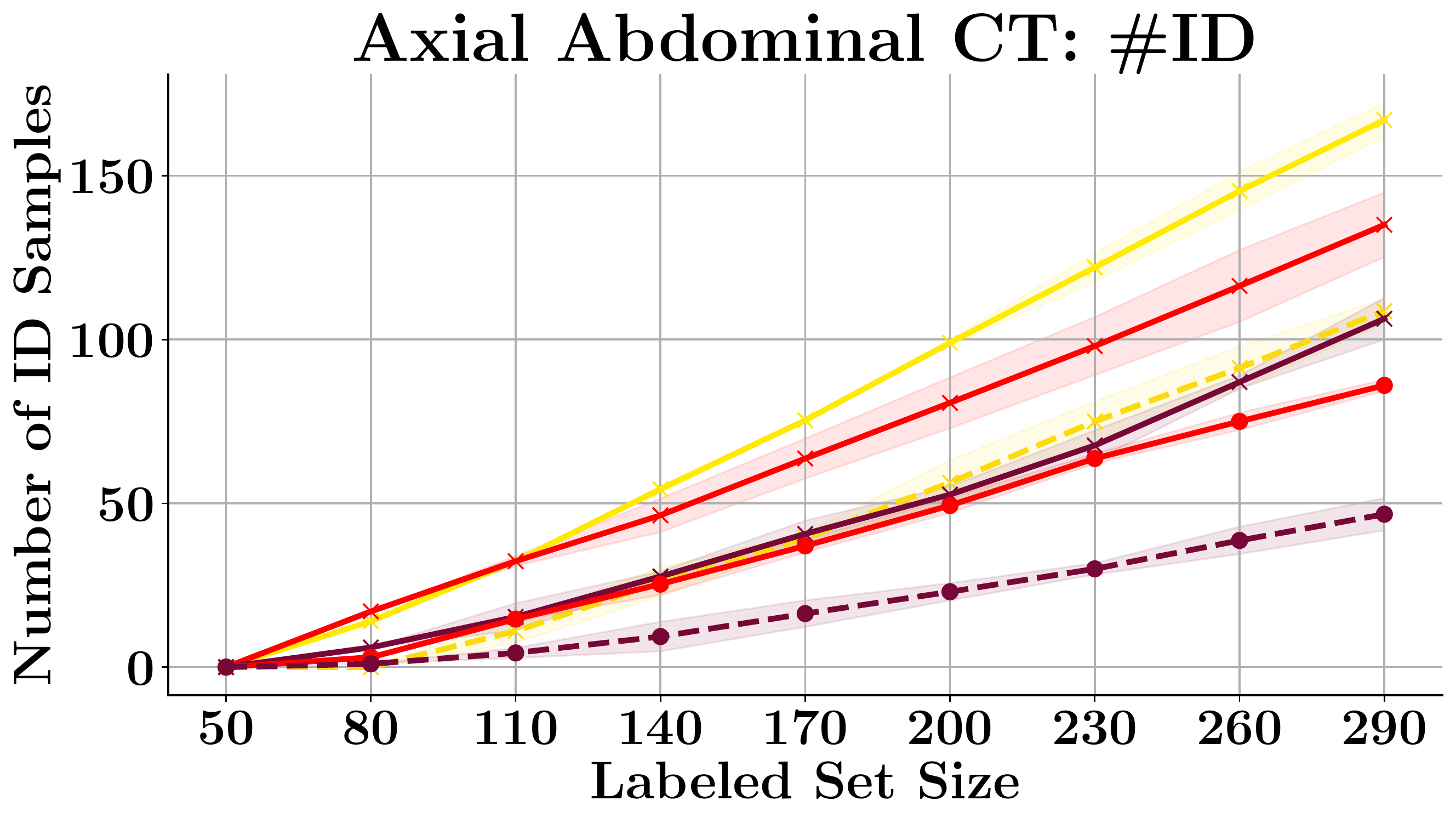}
\end{subfigure}
\caption{Ablation studies comparing the performance of \scmi\ functions with \smi\ functions (\textbf{left} plot) and \scg\ functions (\textbf{right} plot) for scenarios A and C.}
\label{fig:app_ablation_studies}
\end{figure*}

We show the ablation studies for OOD scenarios A and C in \figref{fig:app_ablation_studies}. We observe that jointly modelling similarity and dissimilarity using \scmi\ functions outperforms other methods.

\subsection{Statistical Significance Penalty Matrices} \label{app:pen_matrices}

The penalty matrices computed in this paper follow the strategy used in~\cite{ash2020deep}. In their strategy, a penalty matrix is constructed for each dataset-model pair. Each cell $(i,j)$ of the matrix reflects the fraction of training rounds that AL with selection algorithm $i$ has higher test accuracy than AL with selection algorithm, $j$ with statistical significance. As such, the average difference between the test accuracies of $i$ and $j$ and the standard error of that difference are computed for each training round. A two-tailed $t$-test is then performed for each training round: If $t>t_\alpha$, then $\frac{1}{N_{train}}$ is added to cell $(i,j)$. If $t<-t_\alpha$, then $\frac{1}{N_{train}}$ is added to cell $(j,i)$. Hence, the full penalty matrix gives a holistic understanding of how each selection algorithm compares against the others: A row with mostly high values signals that the associated selection algorithm performs better than the others; however, a column with mostly high values signals that the associated selection algorithm performs worse than the others. As a final note, \cite{ash2020deep} takes an additional step where they consolidate the matrices for each dataset-model pair into one matrix by taking the sum across these matrices, giving a summary of the AL performance for their entire paper that is fairly weighted to each experiment. Below, we present the penalty matrices for each of the settings.

\begin{figure}[!ht]
\centering
\includegraphics[width =0.90\textwidth]{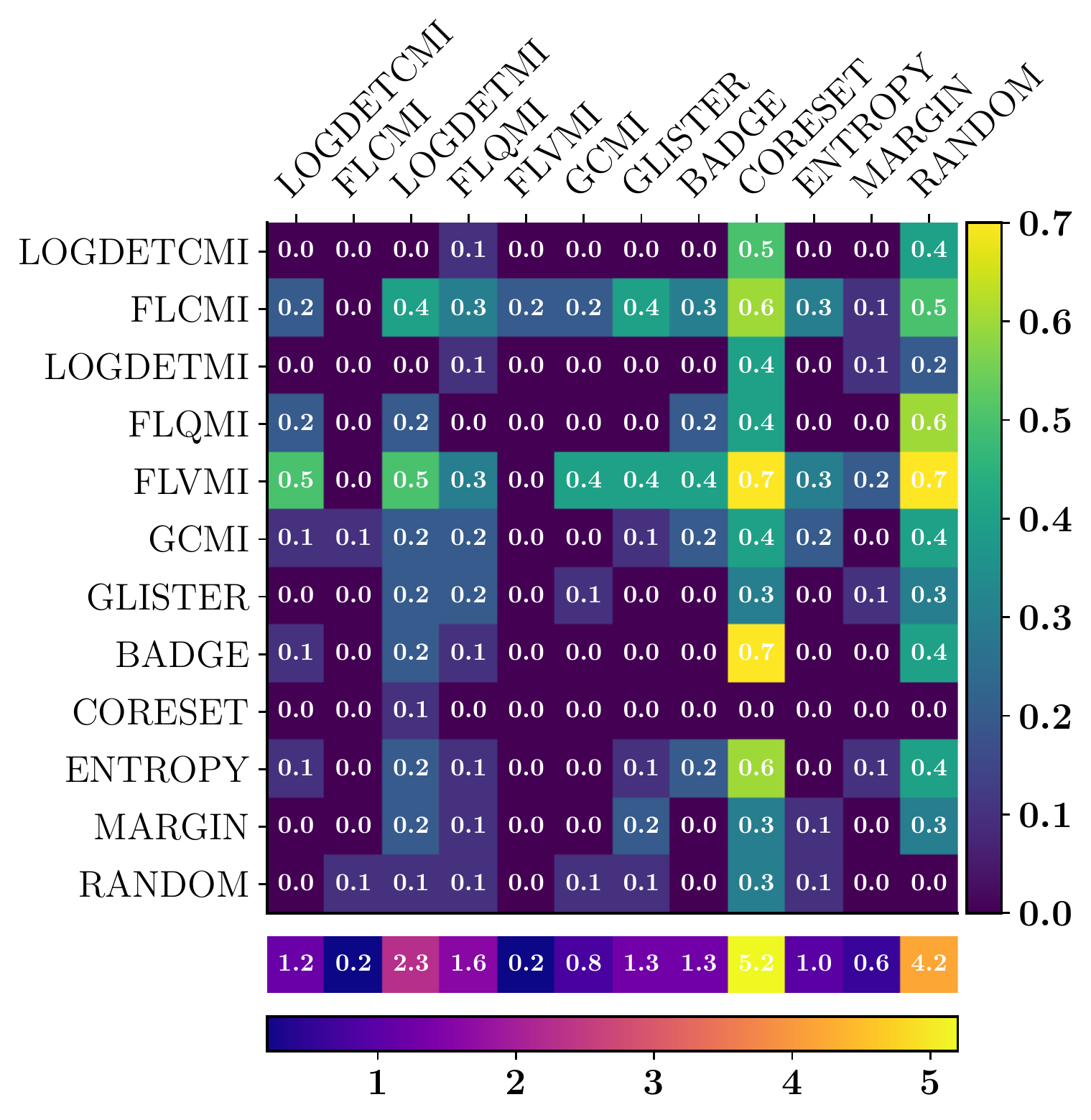}
\caption{
Penalty Matrix for OOD Scenario A comparing the accuracy of active learning across multiple runs. We observe that the SMI functions have a much lower column sum compared to other approaches.
}
\label{pen-matrix1}
\end{figure}

\begin{figure}[!ht]
\centering
\includegraphics[width =0.90\textwidth]{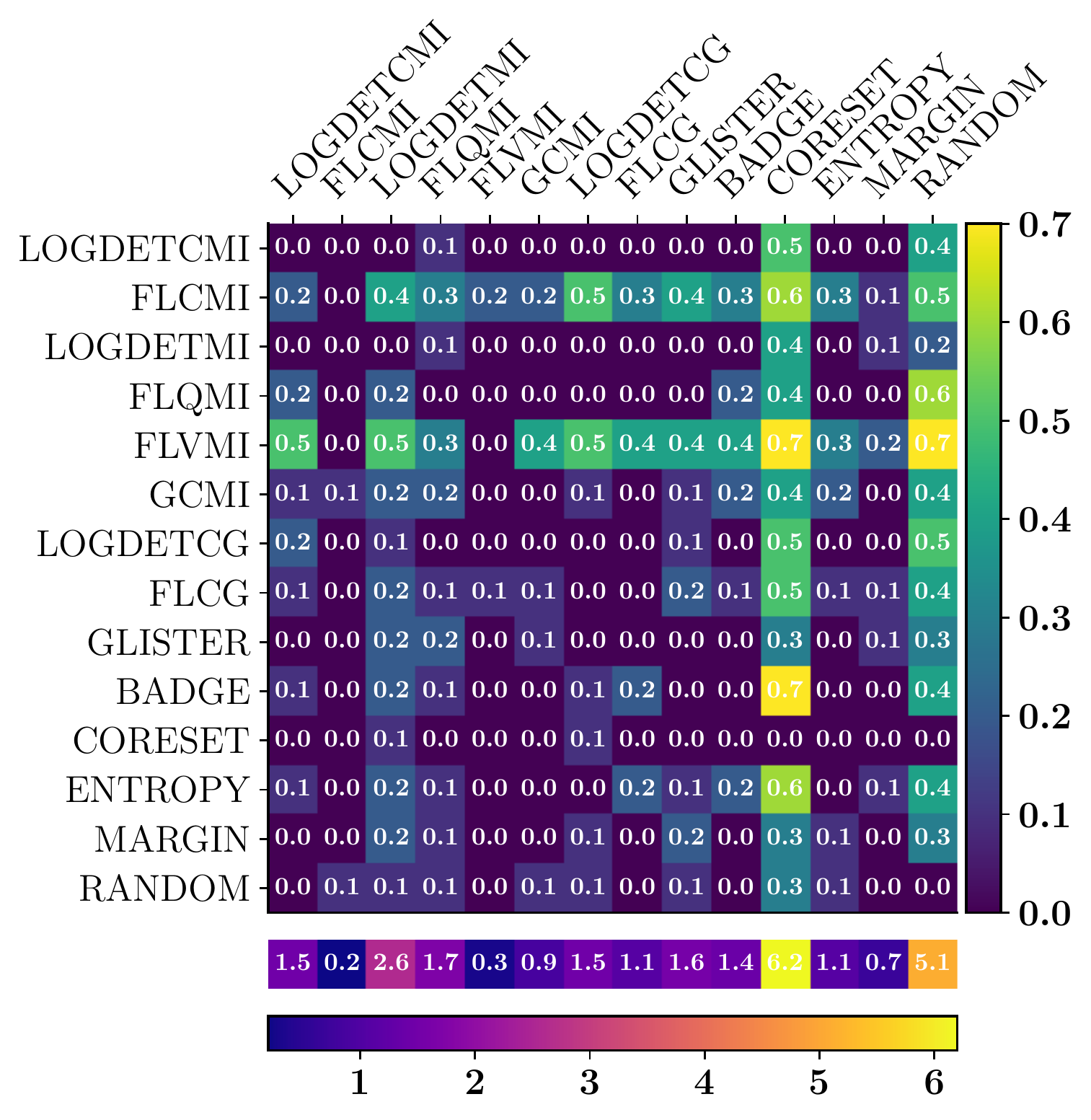}
\caption{
Penalty Matrix for OOD Scenario B comparing the accuracy of active learning across multiple runs. We observe that the SMI functions have a much lower column sum compared to other approaches.
}
\label{pen-matrix2}
\end{figure}

\begin{figure}[!ht]
\centering
\includegraphics[width =0.90\textwidth]{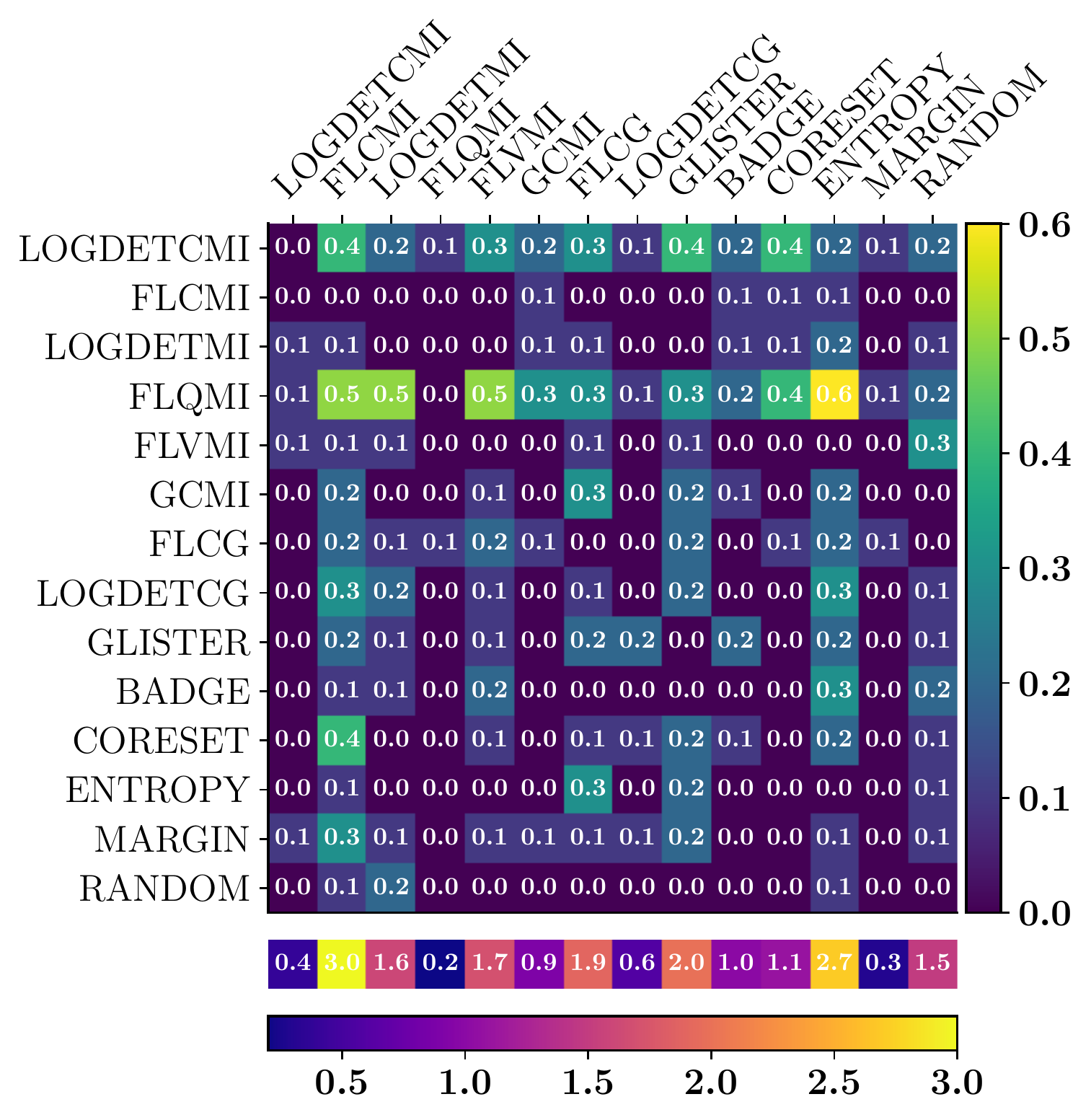}
\caption{
Penalty Matrix for OOD Scenario C comparing the accuracy of active learning across multiple runs. We observe that the SMI functions have a much lower column sum compared to other approaches.
}
\label{pen-matrix3}
\end{figure}

\end{document}